\documentclass[10pt,twocolumn,letterpaper]{article}

\usepackage{cvpr}
\usepackage{times}
\usepackage{epsfig}
\usepackage{graphicx}
\usepackage{amsmath}
\usepackage{amssymb}
\usepackage{algorithm}
\usepackage{algorithmic}
\usepackage{amsmath}
\usepackage{mathrsfs}
\usepackage{multirow}
\usepackage{color}
\usepackage{bm}
\usepackage{float}
\usepackage{booktabs}       
\usepackage{microtype}
\usepackage{subfigure}

\usepackage[breaklinks=true,bookmarks=false]{hyperref}

\cvprfinalcopy 

\def\cvprPaperID{****} 

\begin{document}

\title{Training Meta-Surrogate Model for Transferable Adversarial Attack}

\author{Yunxiao Qin\\
JD Technology\\
{\tt\small qyxqyx@mail.nwpu.edu.cn}
\and
Yuanhao Xiong\\
University of California, Los Angeles\\
{\tt\small yhxiong@cs.ucla.edu}
\and
Jinfeng Yi\\
JD Technology\\
{\tt\small yijinfeng@jd.com}
\and
Cho-Jui Hsieh\\
University of California, Los Angeles\\
{\tt\small chohsieh@cs.ucla.edu}
}

\maketitle

\newcommand{\tabincell}[2]{\begin{tabular}{@{}#1@{}}#2\end{tabular}}  

\begin{abstract}
	We consider adversarial attacks to a black-box model when no queries are allowed. In this setting, many methods directly attack surrogate models and transfer the obtained adversarial examples to fool the target model. 
	Plenty of previous works  investigated what kind of attacks to the surrogate model can generate more transferable adversarial examples, but their performances are still limited due to the mismatches between surrogate models and the target model.
	In this paper, we tackle this problem from a novel angle---instead of using the original surrogate models, can we obtain a {\bf Meta-Surrogate Model} (MSM) such that attacks to this model can be easier transferred to other models?
	We show that this goal can be mathematically formulated as a well-posed (bi-level-like) optimization problem and design a differentiable attacker to make training feasible. 
	Given one or a set of surrogate models, our method can thus obtain an MSM such that adversarial examples generated on MSM enjoy eximious transferability. 
	Comprehensive experiments on Cifar-10 and ImageNet demonstrate that by attacking the MSM, we can obtain stronger transferable adversarial examples to fool black-box models including adversarially trained ones, with much higher success rates than existing methods. The proposed method reveals significant security challenges of deep models and is promising to be served as a state-of-the-art benchmark for evaluating the robustness of deep models in the black-box setting. 
\end{abstract}

\section{Introduction}
\label{introduction}
The developments of Convolutional Neural Network (CNN) \cite{lecun1995convolutional,krizhevsky2012imagenet} have greatly promoted the advancements in Computer Vision~\cite{ren2015faster}.
However, previous works~\cite{goodfellow2014explaining,carlini2017towards,croce2020minimally,ganeshan2019fda} have shown a critical robustness issue that CNN models are vulnerable to human-imperceptible perturbations of input images, also known as adversarial examples (AEs). The design of AEs is useful for revealing the security threats on machine learning systems~\cite{croce2020reliable} and for understanding the representations learned by CNN models~\cite{ilyas2019adversarial}.

In this paper, we consider the problem of black-box attack, where the target victim model is entirely hidden from the attacker. In this setting, standard white-box attacks~\cite{moosavi2016deepfool,carlini2017towards} or even query-based black-box attacks~\cite{ilyas2018black,cheng2018query,cheng2020signopt} cannot be used, and the prevailing way to attack the victim is through transfer attack~\cite{papernot2017practical,wu2018understanding}.  
In transfer attack, the attackers commonly generate AEs by attacking one or an ensemble of {\bf surrogate models} and hope the obtained AEs can also successfully fool the victim black-box model. 

Although great efforts have been made to improve the transferability of adversarial attacks~\cite{tramer2017ensemble, xie2019improving,wu2020skip},  
the transfer attack-based methods still encounter poor success rates, especially when attacking adversarially trained target models.
This is caused by a fundamental limitation of current approaches---they all leverage the surrogate models trained by standard learning tasks (e.g., classification, object detection), while it is not guaranteed that attacks fooling such models can be easily transferred. 
We thus pose the following important question on transfer attack that has not been studied in the literature: Instead of using the original surrogate models, can we obtain a {\bf Meta-Surrogate Model} (MSM) such that attacks to this model can be easier transferred to other models?  

We answer this question in the affirmative by developing a novel black-box attack pipeline called \textbf{\textbf{M}eta-\textbf{T}ransfer \textbf{A}ttack} (MTA). 
Assume a set of source models (standard surrogate models) are given, 
instead of directly attacking these source models, our algorithm aims to obtain a  ``meta-surrogate model (MSM)'', which is designed in the way that attacks to this model can be easier transferred to fool other models, and conduct attacks on the MSM to obtain transferable AEs. 
We show that this goal can be mathematically formulated as a well-posed (bi-level-like) training objective by unrolling the attacks on the MSM and defining a loss to measure the transferability of the resulting AEs. 
To avoid discrete operations in the white-box attack, we propose a Customized PGD attacker that enables back-propagation through the whole procedure. 
With this bi-level-like optimization~\cite{finn2017model,qin2020layer}, the source models supervise the MSM to improve the transferability of the AEs created on it. Through extensive experiments on various models and datasets, we show that the proposed MTA method leads to significantly improved transfer attacks, demonstrating the effectiveness of the MSM.

We summarize the main contributions of our work as follows.
1) We propose a novel MTA framework to train an MSM to improve the transferability of AEs.
To the best of our knowledge, our work is the first attempt to explore a better surrogate model for producing stronger transferable AEs.
2) We compare MTA with state-of-the-art transfer attack methods (M-PGD~\cite{dong2018boosting}, DI-PGD~\cite{xie2019improving}, TI-PGD~\cite{dong2019evading},  SGM-PGD~\cite{wu2020skip}, AEG~\cite{bose2020adversarial}, IR-PGD~\cite{wang2021unified}, SI-N-PGD~\cite{lin2020nesterov}, \emph{etc.}) on Cifar-10~\cite{krizhevsky2009learning} and Imagenet~\cite{deng2009imagenet}. 
The comparisons demonstrate the effectiveness of the proposed MTA---the AEs generated by attacking MSM significantly outperform previous methods,  in attacking  both naturally trained and adversarially trained black-box target models.

\vspace{-6pt}
\section{Background}
\vspace{-3pt}
\label{background}

\textbf{Adversarial attacks.} 
\cite{szegedy2014intriguing} is the earliest work to reveal the interesting phenomenon that CNN models are vunerable to adversarial attacks.
After that, many attacks have been developed~\cite{gao2020patch,zhou2018transferable}. 
Adversarial attacks can be mainly classified into white-box and black-box attacks according to how much information about the target model is exposed to the attacker. 
White-box algorithms~\cite{kurakin2016adversarial} are easier and more effective than black-box algorithms~\cite{brendel2017decision,cheng2018query,cheng2020signopt} to generate adversarial attacks, because they can leverage the full knowledge of the target model including  the model weight, architecture, and gradient.
For example, Fast Gradient Sign Method (FGSM)~\cite{goodfellow2014explaining} uses 1-step gradient ascent to produce AEs that enlarge the model's loss. Projected gradient descent (PGD) attack can be viewed as a multi-step FGSM attack~\cite{madry2017towards}. Many other white-box attacks have also been developed by leveraging full information of the target model~\cite{moosavi2016deepfool,croce2020minimally}. 
In the black-box setting, query-based black-box attacks~\cite{huang2020black,du2020query-efficient} assume 
model information is hidden but attackers can query the model and observe the corresponding hard-label or soft-label predictions. Among them, \cite{chen2017zoo,ilyas2018black} considered soft-label probability predictions and \cite{chen2020hop,huang2020black} considered hard-label decision-based predictions. 
Considering that using a great many of data to query the target model is impractical in many scenes, several researchers proposed ways to further reduce the query counts~\cite{li2020qeba,wang2020spanning}.

\textbf{Transferability of adversarial examples.} In this paper, we consider the black-box attack scenario when the attacker cannot make any query to the target model~\cite{lin2020nesterov,huang2019enhancing,wang2021feature}. 
In this case, the common attack method is based on transfer attack---the attacker generates AEs by attacking one or few surrogate models and hopes the AEs can also fool the target model~\cite{papernot2016transferability,liu2016delving,yuan2021meta,zhou2018transferable}. 
Compared with query-based attacks, crafting AEs from the surrogate model consumes less computational resources and is more realistic in practice. 
Along this direction, subsequent works have made attempts to improve the transferability of AEs. For instance, \cite{dong2018boosting} boosted the transferability by integrating the momentum term into the iterative process. Other techniques like data augmentations~\cite{xie2019improving}, the gradients of skip-connection~\cite{wu2020skip}, and negative interaction between pixels~\cite{wang2021unified} also contribute to stronger transferable attacks. 
In addition to using the original surrogate models, AEG~\cite{bose2020adversarial} adversarially trains a robust classifier together with an encoder-decoder-based transferable perturbation generator.
After the training, AEG uses the generator to generate transferable AEs to attack a set of classifiers.
Compared to all the existing works, our method is the first that meta-trains a new meta-surrogate model (MSM) such that attacks on MSM can be easier transferred to other models. This not only differs from all the previous methods that attack standard surrogate models but also differs from the encoder-decoder based method such as~\cite{bose2020adversarial}.

\vspace{-5pt}
\section{Methodology}
\label{sec:method}
\vspace{-3pt}
We consider the black-box attack setting where the target model is hidden to the attacker and queries are not allowed. 
This setting is also known as the transfer attack  setting \cite{dong2018boosting,dong2019evading,xie2019improving,wang2021unified} and the attacker 1) cannot access the weight, the architecture, and the gradient of the target model; and 2) cannot querying the target model.
The attacker can access 1) the dataset used by  the target model; and 2) a single or a set of {\bf surrogate models} (also known as {\bf source models}) that may share the dataset with the target model. 
For example, it is common to assume that the attacker can access one or multiple well-performing (pretrained) image classification models.
Existing transferable adversarial attack methods conduct various attacks to these models and hope to get transferable AEs that can fool an unknown target model.
Instead of proposing another attack method on surrogate models, we propose a novel framework MTA to train a {\bf Meta-Surrogate Model (MSM)} with the goal that attacking the MSM can generate stronger transferable AEs than directly attacking the original surrogate models. 
When evaluating, the transferable AEs are generated by attacking the MSM with standard white-box attack methods (e.g., PGD attack). In the following, we will first review exiting attacks and then show how to form a bi-level optimization objective to train the MSM model.

\vspace{-3pt}
\subsection{Reviews of FGSM and PGD}
\vspace{-3pt}
We follow the settings of existing works~\cite{dong2018boosting,xie2019improving,wu2020skip,wang2021unified} to focus on untargeted attack, where the attack is considered successful as long as the perturbed image is wrongly predicted.

\noindent \textbf{FGSM}~\cite{goodfellow2014explaining} conducts one-step gradient ascent to generate AEs to enlarge the prediction loss.
The formulation can be written as
\begin{equation}
x_{adv} = \text{Clip}\big(x + \epsilon \cdot \text{sign}(\nabla_{x} \emph{L}(f(x), y ))\big),
\label{eq:FGSM}
\end{equation}
where $x$ is a clean image and $y$ is the corresponding label; 
$\epsilon$ is the attack step size that determines the maximum $L_\infty$ perturbation of each pixel; 
$f$ is the victim model that is transparent to the FGSM attacker; 
$\text{Clip}$ is the function that clipping the values of $x_{adv}$ to the legal range (\emph{e.g.}, clipping the RGB AEs to the range of $[0, 255]$); $L$ is usually the cross-entropy loss.

\noindent \textbf{PGD}~\cite{kurakin2016adversarial}, also known as I-FGSM attack,
is a multi-step extension of FGSM. 
The formulation of PGD is
\begin{equation}
x_{adv}^{k} \!=\! \text{Clip}\big(x_{adv}^{k-1} \!+\! \frac{\epsilon}{T} \cdot \text{sign}(\nabla_{x_{adv}^{k-1}} \emph{L}(f(x_{adv}^{k-1}), y) )\big).
\label{eq:PGD}
\end{equation}
$x_{adv}^{k}$ is the AEs generated in the $k$-th gradient ascent step.
Note that $x_{adv}^{0}$ is the clean image equals to $x$.
Eq~\eqref{eq:PGD} will be run for $T$ iterations to obtain  $x_{adv}^{T}$ with perturbation  size $\epsilon$.

\begin{figure}[t]
	\centering
	\includegraphics[width=0.49\textwidth]{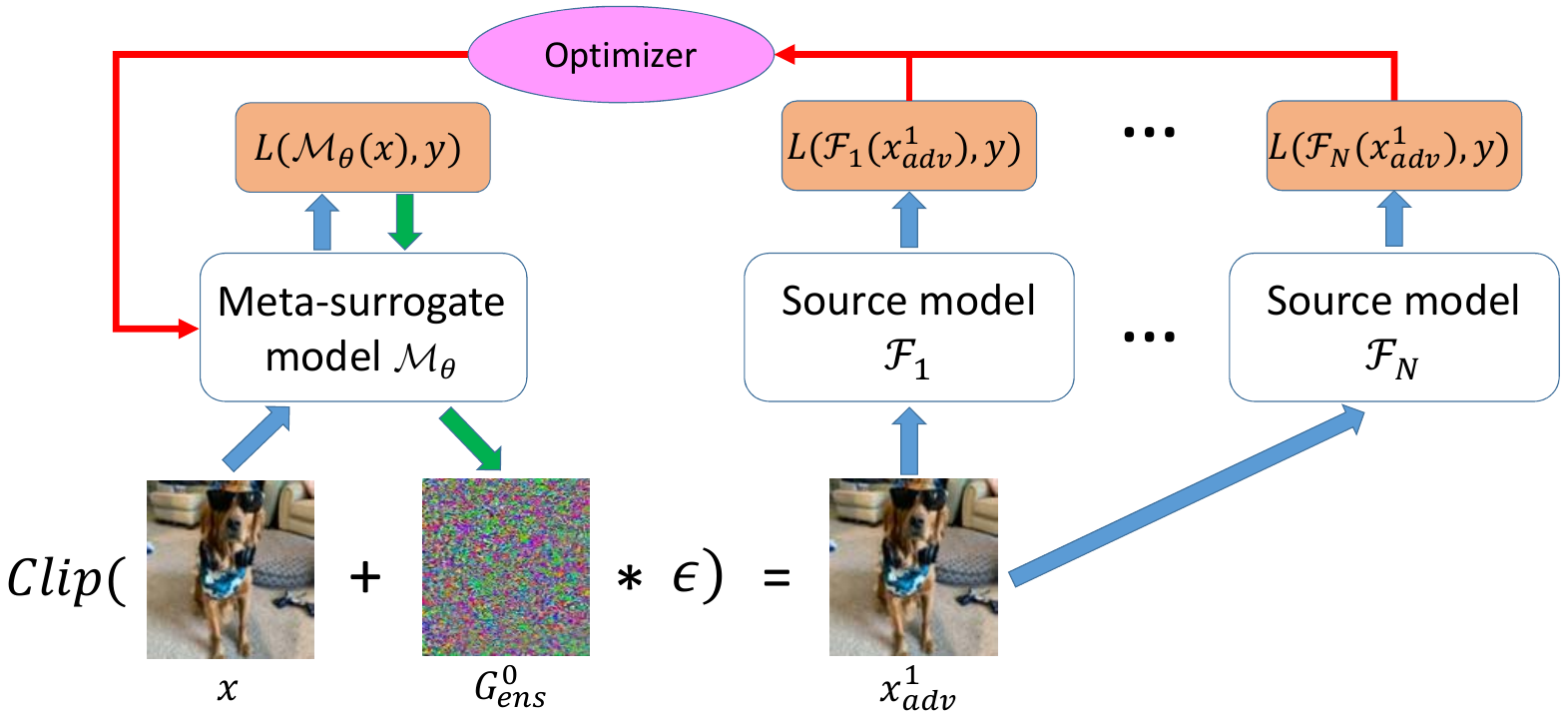}
	\caption{
		The framework of the proposed MTA when $T=1$ and $\mathcal{A}(\mathcal{M}_\theta(x)) = x_{adv}^1$.
		The clean image $x$ is first feed into the MSM $\mathcal{M}_\theta$ and obtain the loss $\emph{L}(\mathcal{M}_\theta(x),y)$.
		Next we back-propagate the loss and use Eq~\eqref{eq:gradient_ensemble} to obtain the noise $g_{ens}^0$.
		Then, via Eq~\eqref{eq:my_PGD}, we obtain the adversarial example $x_{adv}^1$ which will be feed into the source models $\mathcal{F}_1$, $\mathcal{F}_2$, ..., and $\mathcal{F}_N$.
		Finally, by maximizing the source models' loss, we can optimize the MSM to learn a particular weight so that the adversarial example $x_{adv}^1$ attacking it can fool source models.
	}
	\label{fig:framework}
	\vspace{-8pt}
\end{figure}

\vspace{-4pt}
\subsection{Meta-transfer attack}
\vspace{-1pt}
How to train the MSM where attacks to this model can be easier transferred to other models? We show this can be formulated as a bi-level training objective.  Let $\mathcal{A}$ denote an attack algorithm (e.g., FGSM or PGD) and $\mathcal{M_{\theta}}$ denote the {\bf MSM} parameterized by $\theta$.
For a given image $x$, the AE generated by attacking $\mathcal{M_{\theta}}$ can be denoted as $\mathcal{A}(\mathcal{M}_\theta, x, y)$. 
For example, if $\mathcal{A}$ is FGSM, then $\mathcal{A}(\mathcal{M}_\theta, x, y) = x_{adv} = \text{Clip}\big(x + \epsilon \cdot \text{sign}(\nabla_{x} \emph{L}(\mathcal{M_{\theta}}(x), y ))\big)$.
Since in the attack time we only have access to a set of source models $\mathcal{F}_1, \dots, \mathcal{F}_N$, 
we can evaluate the transferability of the adversarial example $\mathcal{A}(\mathcal{M}_\theta, x, y)$ on the source models and optimize the MSM via maximizing the adversarial losses of those $N$ source models, leading to the following training objective: 
\begin{equation}
\arg\max_{\theta} \mathbb{E}_{(x, y)\sim D} \big[ \textstyle \sum_{i=1}^N  L(\mathcal{F}_i(\mathcal{A}(\mathcal{M}_\theta, x, y)), y) \big], 
\label{eq:meta_attack}
\end{equation}
where $D$ is the distribution of training data. 
The structure of this objective and the training procedure can be illustrated in Figure~\ref{fig:framework}, where we can view it as a meta-learning or bi-level optimization method. 
At the lower level, the AE is generated by a white-box attack (usually gradient ascent) on MSM, while at the higher level, we feed the AE to the source models to compute the robust loss. 
Solving Eq~\eqref{eq:meta_attack} will find an MSM where attacking it leads to stronger transferable AEs. 
The optimization steps of Eq~\eqref{eq:meta_attack} are detailed below.

\textbf{First}, $\mathcal{A}$ should be some strong white-box attacks, such as FGSM or PGD.
However, directly using those attacks will make the gradient of meta training objective Eq~\eqref{eq:meta_attack} ill-defined since the $\text{sign}$ function in both FGSM and PGD introduce a discrete operation.
This results in that the gradient back-propagating through $\text{sign}$ be zero and further prohibits the training of the MSM.

To overcome this challenge, we design $\mathcal{A}$ as an approximation of PGD and denote it as Customized PGD.
Section~\ref{sec:analyse} will show more explanation about how the $\text{sign}$ function in PGD prohibits back-propagation and how Customized PGD enables the back-propagation.
The crucial difference between PGD and the Customized PGD is the operation to the gradient $\nabla_{x_{adv}^{k-1}} \emph{L}(\mathcal{M}_\theta(x_{adv}^{k-1}), y)$, where $L$ is the cross entropy loss.
For simplicity, we denote the vanilla gradient $\nabla_{x_{adv}^{k}} \emph{L}(\mathcal{M}_\theta(x_{adv}^{k}), y)$ at the $k$-th step as $g^k$, and generate another map $g_{ens}^k$ via Eq~\eqref{eq:gradient_ensemble}: 
\begin{equation}
\left\{
\begin{array}{lr}
g_{1}^k = \frac{g^k}{\text{sum}(\text{abs}(g^k))}   \\
g_{t}^k =  \frac{2}{\pi} \cdot  \text{arctan}(\frac{g^k}{\text{mean}(\text{abs}(g^k))})  \\
g_{s}^k =  \text{sign}(g^k) \\
g_{ens}^k = g_{1}^k + \gamma_1 \cdot g_{t}^k + \gamma_2 \cdot g_{s}^k
\end{array}
\right. 
\label{eq:gradient_ensemble}
\end{equation}
Note that we set $\gamma_1=\gamma_2=0.01$ as default for all the experiments.
Both $g_{1}^k$ and $g_{t}^k$ ensure the objective in Eq~\eqref{eq:meta_attack} be differentiable with respect to the MSM's weight $\theta$.
$\text{arctan}$ is a smooth approximation of sign and $\frac{1}{\text{mean}(\text{abs}(g^k))}$ prevents arctan from falling into the saturation or linear region. 
The item $\gamma_2 \cdot g_{s}^k$ provides the lower-bound for each pixel's perturbation in $g_{ens}^k$.
The experiments in Section~\ref{sec:ablation_gamma} will demonstrate the importances of $g_{t}^k$ and $g_{s}^k$ for Customized PGD.
With Eq~\eqref{eq:gradient_ensemble}, the Customized PGD conducts the following update to generate AE: 
\begin{equation}
x_{adv}^{k} = \text{Clip}(x_{adv}^{k-1} + \frac{\epsilon_c}{T} \cdot g_{ens}^{k-1}).
\label{eq:my_PGD}
\end{equation}
Note that  $\epsilon_c$ differs from the perturbation $\epsilon$ in FGSM and PGD because $g^{k-1}_{ens}$ in our update is not a sign vector and its size will depend on the magnitude of the original gradient. 
Finally, we get $x_{adv}^{T}$ after $T$ iterations of Eq~\eqref{eq:my_PGD}.

\textbf{Second}, we feed $x_{adv}^{T}$ into $N$ source models and calculate the corresponding adversarial losses $\emph{L}(\mathcal{F}_i(x_{adv}^{T}), y)$ for all $i=1, \dots, N$. 
Larger losses of the $N$ source models indicate a higher likelihood that $x_{adv}^T$ fooling the MSM can also fool the source models.

\textbf{Third}, we optimize the MSM by maximizing the objective function defined in Eq~\eqref{eq:meta_attack}. The update rule can be written as
\begin{equation}
\theta^{'} = \theta + \alpha \cdot
\textstyle \sum_{i=1}^N
\nabla_{\theta} 
\emph{L}(\mathcal{F}_i (x_{adv}^{T}), y), 
\label{eq:optimize}
\end{equation}
where $x_{adv}^T$ can be written as a function of $\theta$ by unrolling the attack update rule Eq~\eqref{eq:my_PGD} $T$ times. We will show how to explicitly compute the gradient in Section~\ref{sec:analyse}. 
With this training procedure, the MSM is trained to learn a particular weight with which the white-box AEs fooling it can also fool other models. 
We summarize the training and testing of MTA in Algorithm~\ref{algorithm:train} and Appendix, respectively.
Each capitalized notation represents a batch of the variable denoted with lower case.
For example, $X$ denotes a batch of $x$.
Note that Customized PGD is just a continuous approximation of PGD used to train the MSM. In the inference phase, we use standard attacks such as PGD to craft AEs on the MSM.

\vspace{-3pt}
\subsection{Gradient calculation}
\vspace{-3pt}
\label{sec:analyse}
In the calculation we set both $N$ and $T$ in Eq~\eqref{eq:optimize} to $1$, so the gradient
in Eq~\eqref{eq:optimize} is $\nabla_{\theta} \emph{L}(\mathcal{F}_1(x_{adv}^{1}), y)$. 
According to Eq~\eqref{eq:my_PGD}, we can replace $x_{adv}^{1}$ in Eq~\eqref{eq:optimize} with $\text{Clip}(x_{adv}^{0} + \epsilon_c \cdot g_{ens}^{0})$, where $x_{adv}^{0}$ equals to $x$.
For simplicity, we ignore the clip function in the analysis and simplify the derivation as $\nabla_{\theta} \emph{L}(\mathcal{F}_1(x + \epsilon_c \cdot g_{ens}^{0}), y)$. By chain rule and since $x$ is independent to $\theta$, we can further rewrite this as 
\begin{equation}
\frac{\partial \emph{L}(\mathcal{F}_1(x+\epsilon_c \cdot g_{ens}^0), y)}{\partial g^0_{ens}} \cdot \frac{\partial g^0_{ens}}{\partial \theta}.
\label{eq:optimize1}
\end{equation}
By replacing $g_{ens}^{0}$ with Eq~\eqref{eq:gradient_ensemble}, the second term of Eq~\eqref{eq:optimize1} can be expanded as
	\begin{equation}
	\nabla_{\theta}g_{ens}^{0} = \nabla_{\theta}g_{1}^0 + \gamma_1 \cdot \nabla_{\theta}g_{t}^0 + \gamma_2 \cdot \nabla_{\theta}g_{s}^0.
	\label{eq:optimize2}
	\end{equation}
Note that $g_{s}^0$ equals to $\text{sign}(g^0)$ and the sign function introduces discrete operation so that the gradient of $g_{s}^0$ with respect to $\theta$ becomes 0 (unless $g^0=0$).
Therefore, $\nabla_{\theta}g_{ens}^{0}$ can be further written as
	\begin{equation}
	\begin{aligned}
	\nabla_{\theta}g_{ens}^{0} &= \nabla_{\theta}g_{1}^0 + \gamma_1 \cdot \nabla_{\theta}g_{t}^0 \\
	&= \nabla_{\theta}(\frac{\nabla_{x} \emph{L}(\mathcal{M}_\theta(x), y)}{\text{sum}(\text{abs}(\nabla_{x} \emph{L}(\mathcal{M}_\theta(x), Y)))})  \\
	&+ \gamma_1 \cdot \nabla_{\theta}(\text{arctan}(\frac{\nabla_{x} \emph{L}(\mathcal{M}_\theta(x), y)}{\text{mean}(\text{abs}(\nabla_{x} \emph{L}(\mathcal{M}_\theta(x), y)))})).
	\end{aligned}
	\label{eq:optimize3}
	\end{equation}
In this formulation, $\nabla_{x} \emph{L}(\mathcal{M}_\theta(x), y)$ depends on $\theta$ and the second-order derivative of $\nabla_{x} \emph{L}(\mathcal{M}_\theta(x), y)$ \emph{w.r.t} $\theta$ can be obtained with lots of deep learning libraries~\cite{abadi2016tensorflow,paszke2017automatic}.
In summary,  by integrating Eqs.\eqref{eq:optimize}-\eqref{eq:optimize3}, the MSM can be optimized by an SGD-based  optimizer.

\begin{algorithm}[t]
	\caption{Training of Meta-Transfer Attack}\label{algorithm:train}
	{\bfseries input:} $N$ source models $\mathcal{F}_1, \dots, \mathcal{F}_N$, Training set $\mathbb{D}$, batch size $b$, initialized MSM $\mathcal{M}_\theta$. \\
	{\bfseries output:} Optimized weight $\theta$. \\
	{\bfseries 1 \;\!:} {\bfseries while not} done {\bfseries do} \\
	{\bfseries 2 \;\!:} 	\quad	sample data ($X$=$[x_1,\dots,x_b]$, $Y$=$[y_1,\dots, y_b]$) $\in \mathbb{D}$\\
	{\bfseries 3 \;\!:} 	\quad	$X_{adv}^{0} = X$ \\
	{\bfseries 4 \;\!:} 	\quad	\textbf{for} k in [1, 2, ..., T]: \\
	{\bfseries 5 \;\!:}     \quad\quad $G^k = \nabla_{X_{adv}^{k-1}} \emph{L}(\mathcal{M}_\theta(X_{adv}^{k-1}), Y)$ \\
	{\bfseries 6 \;\!:}     \quad\quad obtain $G^k_{ens}$ via Eq~\eqref{eq:gradient_ensemble} \\
	{\bfseries 7 \;\!:}     \quad\quad obtain $X_{adv}^{k}$ via Eq~\eqref{eq:my_PGD}\\
	{\bfseries 8 \;\!:}     \quad \textbf{end for} \\
	{\bfseries 9 \;\!:}  \quad   {\bfseries for} each source model $\mathcal{F}_i \in [\mathcal{F}_1, \mathcal{F}_2, \dots, \mathcal{F}_N]$, \ \ {\bfseries do} \\
	{\bfseries 10:} \qquad	evaluate $X_{adv}^{T}$ on $\mathcal{F}_i$ and obtain $\emph{L}(\mathcal{F}_i(X_{adv}^{T}), Y)$   \\
	{\bfseries 11:} 	\quad \textbf{end for} \\
	{\bfseries 12:} 	\quad  $\theta = \theta + \alpha \cdot \nabla_{\theta} \textstyle{\sum_{i}^{N}} \emph{L}(\mathcal{F}_i(X_{adv}^{T}), Y)$\\
	{\bfseries 13:} 	{\bfseries return} $\theta$
\end{algorithm}

\vspace{-5pt}
\section{Experimental Results}
\vspace{-3pt}
\label{experiment}
We conduct experiments to show that the proposed method, under the same set of source models, can generate stronger transferable AEs than existing transfer attack methods. 

We first present our general experimental settings.
\textbf{1)} We conduct experiments on both Cifar-10~\cite{krizhevsky2009learning} and ImageNet~\cite{deng2009imagenet}.
\textbf{2)} We compare the proposed MTA with seven state-of-the-art transferable adversarial attack methods, including M-PGD~\cite{dong2018boosting}, DI-PGD~\cite{xie2019improving}, TI-PGD~\cite{dong2019evading}, SGM-PGD~\cite{wu2020skip}, SI-N-PGD~\cite{lin2020nesterov}, AEG~\cite{bose2020adversarial}, IR-PGD~\cite{wang2021unified}, and FIA~\cite{wang2021feature}. 
Note that since SGM-PGD is based on enlarging the gradient of skip connections, we only include this method on ImageNet experiments when the source models have sufficient skip connections. 
The baseline AEG is compared only on Cifar-10 because the official AEG is evaluated only on small scale dataset (Mnist and Cifar-10) and is computation costly to train the perturbation generator on large-scale dataset.
\textbf{3)} Since the number of attack iterations $T$ is set differently in training and testing, we denote it as $T_t$ in training and $T_v$ in evaluation, respectively, to avoid confusion. 
\textbf{4)} When training the MSM, we use the Customized PGD with $\gamma_1$=$\gamma_2$=0.01 to attack the MSM.
When evaluating, we use PGD with $T_v$=10 and $\epsilon$=15 to attack the MSM and denote the generated adversarial attacks as MTA-PGD.
\textbf{5)} When using the baseline methods to generate AEs on multiple source models, we follow \cite{dong2018boosting} to ensemble the logits of the source models before loss calculation.
\textbf{6)} We use source and target models to train and to evaluate the MSM, respectively.
\textbf{7)} 
For fair comparisons between MTA-PGD and baselines, we implement baselines with the number of iterations $T$=10 and $\epsilon$=15, and other hyper-parameters are tuned for their best possible performances (implementations are detailed in Appendix).
\textbf{8)} More experiments (\emph{e.g.}, targeted transfer attack, attacks with smaller $\epsilon$, more comparisons between MTA and baselines) will be shown in Appendix.

\begin{table*}[t]
	\centering
	\caption{The transfer attack success rates on eight target networks on Cifar-10.
		The MSM is trained with eight source models. 
		From left to right, the eight target models are MobileNet-V3 (MN-V3), ShuffleNet-V1 (SN-V1), -V2 (SN-V2), SqueezeNet-A (SN-A), -B (SN-B), and adversarially trained ResNet-18 (Res-18$_{adv}$), ResNet-34 (Res-34$_{adv}$), and SeResNet-50 (SeRes-50$_{adv}$).
	}
	{\small 
		\begin{tabular}{ccccccccc}
			\toprule
			Method  & MN-V3 & SN-V1 & SN-V2 & SN-A & SN-B &  Res-18$_{adv}$ & Res-34$_{adv}$ & SE-50$_{adv}$ \\
			\hline
			PGD~\cite{kurakin2016adversarial}  		 & 51.8\% & 64.1\% & 49.4\% & 57.2\% & 56.3\% & 67.7\% & 63.9\% & 63.4\% \\
			DI-PGD~\cite{xie2019improving}  		 & 57.8\% & 72.5\% & 56.4\% & 65.7\% & 64.6\% & 80.7\% & 73.1\% & 71.0\% \\
			M-PGD~\cite{dong2018boosting}  			& 70.2\% & 85.6\% & 72.6\% & 83.7\% & 83.0\% & 92.9\% & 90.9\% & 89.1\% \\
			A-PGD~\cite{croce2020reliable}  			& 74.1\% & 88.9\% & 75.8\% & 84.2\% & 83.6\% & 90.7\% & 89.3\% & 89.1\% \\
			TI-PGD~\cite{dong2019evading}  			& 54.5\% & 59.9\% & 54.2\% & 71.8\% & 71.4\% & 57.6\% & 46.3\% & 46.6\% \\
			AEG~\cite{bose2020adversarial}  			& 90.8\% & 92.5\% & 85.8\% & 91.3\% & 91.0\% & 96.1\% & 93.6\% & 93.1\%  \\
			IR-PGD~\cite{wang2021unified}  			& 59.3\% & 77.9\% & 62.5\% & 71.6\% & 69.1\% & 79.8\% & 73.7\% & 72.1\%  \\
			\textbf{MTA-PGD }  	  &  \textbf{91.8\%} &  \textbf{98.4\%} & \textbf{90.9\%} & \textbf{94.9\%} & \textbf{93.8\%} & \textbf{98.4\%} & \textbf{96.5\%} & \textbf{97.1\%}  \\
			\textbf{MTA-PGD$_{\gamma_1=0}$ }  	  &  70.0\% &  {80.9\%} & {68.5\%} & {58.5\%} & 59.4\%  & 67.7\% & 59.2\% & 68.9\%  \\
			\textbf{MTA-PGD$_{\gamma_2=0}$ }  	  &  {90.0\%} &  {98.2\%} & {90.5\%} & {93.9\%} & 93.1\% &97.6\% & 96.0\% & 96.3\%  \\
			\textbf{MTA-PGD$_{dense}$ }  	  &  86.9\% &  96.2\% & {87.1\%} & {89.0\%} & 87.6\%  & 96.2\% & 91.3\% & 93.6\%  \\
			\bottomrule
		\end{tabular}
	}
	\label{tab:cifar}
\end{table*}

\begin{figure*}[h!]
	\centering
	\includegraphics[width=0.8\textwidth]{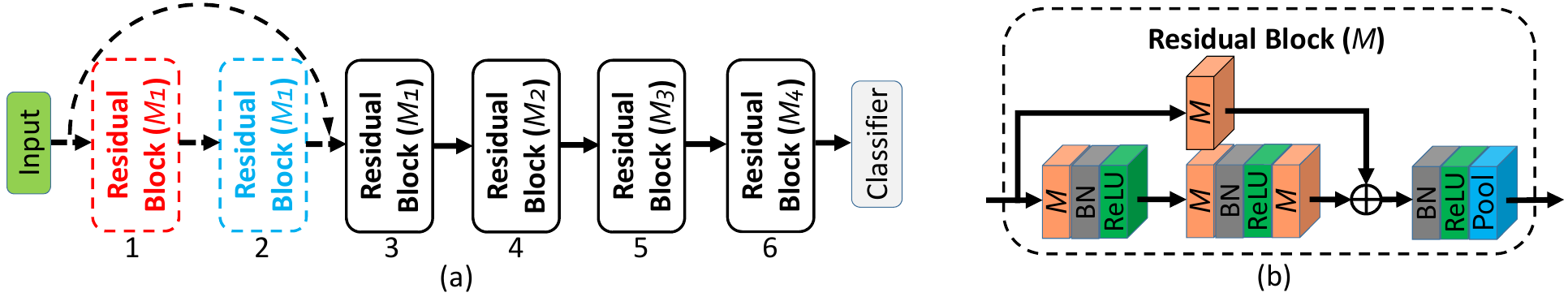}
	\caption{
		(a) The structures of ResNet-13 and -19.
		ResNet-13 contains the top four blocks in the solid-line box and the classifier.
		ResNet-19 contains all the six blocks and the classifier.
		The parameter $M*$ of each block denotes the number of filters of its convolution layers.
		(b) The detailed structure of residual block.
		The orange cube is the convolution layer and the number on it denotes its number of filters.
		Pool in the sixth block is global-average pooling while all the other pool is max-pooling with both stride and kernel size of 2$\times$2.
		The convolution layer in the shortcut path uses 1$\times$1 kernel size while all the other convolution layers use 3$\times$3.
	}
	\label{fig:network}
\end{figure*}

\vspace{-5pt}
\subsection{Experiments on Cifar-10}
\vspace{-3pt}
\subsubsection{Experimental configurations}
\vspace{-3pt}
\label{sec:exp_set_cifar}
On Cifar-10, we use 8 source models including ResNet-10, -18, -34~\cite{he2016deep}, SeResNet-14, -26, -50~\cite{hu2018squeeze}, MobileNet-V1~\cite{howard2017mobilenets}, and -V2~\cite{sandler2018mobilenetv2} to train the MSM.
To ensure mismatches between the source and target models and to avoid saturated transfer attack performances (\emph{i.e.}, attack success rates close to 100\%), we select the 8 target models including MobileNet-V3~\cite{howard2019searching}, ShuffleNet-V1, -V2~\cite{zhang2018shufflenet}, SqueezeNet-A, -B~\cite{iandola2016squeezenet}, and adversarially trained ResNet-18, -34, and SeResNet-50.
The network architectures of all 16 models are defined on public  GitHub repositories\footnote{https://github.com/yxlijun/cifar-tensorflow}$^,$\footnote{https://github.com/TropComplique/ShuffleNet-tensorflow}$^,$\footnote{https://github.com/TropComplique/shufflenet-v2-tensorflow}.
We train all the other 16 models and describe the training details of these models in Appendix.
The trained models and the code will be released to the community for reproducibility.

\noindent \textbf{Training the MSM}.
The default network architecture of the MSM is ResNet-13 shown in Figure \ref{fig:network}, with $M1$, $M2$, $M3$, and $M4$ set to 64, 128, 256, and 512, respectively.
We use the 8 source models to train the MSM for 60 epochs with the number of attack steps $T_t$ of 7.
$\epsilon_c$ of the Customized PGD is initialized to 1,600 and is exponentially decayed by 0.9$\times$ for every 4,000 iterations.
The learning rate $\alpha$ and the batch size are set to 0.001 and 64, respectively.

\noindent \textbf{Evaluating the MSM}.
On each target model, we only attack the correctly classified test images because attacking wrongly classified clean images is less meaningful.

\begin{table*}[t]
	\centering
	\caption{The transfer attack success rates on seven black-box networks where we use one source model.
	}
	{\small 
		\begin{tabular}{ccccccccc}
			\toprule
			Source & Method & Inc-V3 & Inc-V4 & IncRes-V2 & Res-152 & Inc-V3$_{_{ens3}}$ & Inc-V3$_{_{ens4}}$ & IncRes-V2$_{_{ens}}$ \\
			\hline
			\multirow{6}{*}{Inc-V3}
			&DI-PGD~\cite{xie2019improving}  	     &/   &35.2\% & 28.2\% & 22.3\% & 5.1\% & 4.3\% & 2.5\% \\
			&M-PGD~\cite{dong2018boosting}  			  &/  &38.1\% & 35.8\% & 29.6\% & 9.1\% & 8.8\% & 4.5\% \\
			&M-DI-PGD      &/   &61.7\% & 57.3\% & 48.0\% & 13.6\% & 12.0\% & 6.5\% \\
			&SI-N-PGD~\cite{lin2020nesterov}      &/   &63.8\% & 62.0\% & 51.7\% & 25.5\% & 25.2\% & 12.4\% \\
			&IR-PGD~\cite{wang2021unified}      &/   &33.6\% & 28.1\% & 15.9\% & 5.1\% &5.5\% & 3.0\% \\
			&FIA~\cite{wang2021feature}      &/   &69.0\% & 66.8\% & 52.5\% & 29.3\% &27.7\% & 14.9\% \\
			&\textbf{MTA-PGD}  	 &  /  & {90.9\%} &  {87.3\%} &  {74.1\%} &  {67.7\%} &  {39.3\%} & {26.1\%} \\
			&\textbf{MTA-IR-PGD}  	 &  /  & \textbf{95.5\%} &  \textbf{93.2\%} &  \textbf{85.0\%} &  \textbf{83.5\%} &  \textbf{56.9\%} & \textbf{40.7\%} \\
			\hline
			\multirow{6}{*}{Inc-V4}
			&DI-PGD~\cite{xie2019improving}  		    &44.9\%   &/ & 30.5\% & 26.7\% & 5.9\% & 5.5\% & 3.3\% \\
			&M-PGD~\cite{dong2018boosting}  			    &52.7\%  &/ & 41.8\% & 37.3\% & 12.4\% & 11.0\% & 5.8\% \\
			&M-DI-PGD  	  &69.1\%   &/ & 58.7\% & 49.3\% & 16.6\% & 14.1\% & 8.2\% \\
			&SI-N-PGD~\cite{lin2020nesterov}      &74.6\%   &/ & 67.3\% & 61.6\% & 39.2\% & 35.9\% & 22.0\% \\
			&IR-PGD~\cite{wang2021unified}      &46.5\%   & /& 33.2\% & 18.9\% & 8.1\% & 8.8\% & 4.9\% \\
			&FIA~\cite{wang2021feature}      &63.6\%   &/ & 55.2\% & 45.9\% & 28.5\% &26.1\% & 16.8\% \\
			&\textbf{MTA-PGD}  &  {87.3\%}  & / &  {84.7\%} &  {73.1\%} &  {61.7\%} &  {38.2\%} & {29.0\%} \\
			&\textbf{MTA-IR-PGD}  &  \textbf{93.3\%}  & / &  \textbf{90.5\%} &  \textbf{82.0\%} &  \textbf{77.2\%} &  \textbf{57.7\%} & \textbf{44.9\%} \\
			\hline
			\multirow{7}{*}{IncRes-V2}
			&DI-PGD~\cite{xie2019improving}  		   &46.9\%   &42.0\% & / & 29.5\% & 8.6\% & 6.5\% & 5.5\% \\
			&M-PGD~\cite{dong2018boosting}  				&53.2\%  &45.2\% & / & 38.8\% & 16.2\% & 13.3\% & 9.7\% \\
			&M-DI-PGD  	  &{64.7\%}   &{61.7\%} & / & 50.6\% & 23.7\% & 18.6\% & 13.6\% \\
			&SI-N-PGD~\cite{lin2020nesterov}      &\textbf{78.2\%}   &\textbf{70.7\%} & / & 63.8\% & 45.2\% & 38.8\% & \textbf{32.9\%}  \\
			&IR-PGD~\cite{wang2021unified}      &49.7\%   &44.9\% & / & 25.2\% & 13.6\% & 11.2\% & 10.9\% \\
			&FIA~\cite{wang2021feature}      &63.2\%   &57.8\% & / & 51.3\% & 35.1\% &30.3\% & 25.0\% \\
			&\textbf{MTA-PGD}    &  44.7\%  & 41.7\% &  / &  57.9\%&  23.5\%&  19.4\%& 17.5\% \\
			&\textbf{MTA-PGD$_{Inc}$}    &  64.3\%  & 51.7\% &  / &  {76.0\%} &  {46.2\%} &  {39.3\%} & {27.5\%} \\
			&\textbf{MTA-IR-PGD$_{Inc}$}    &  {66.2\%}  & 52.3\% &  / &  \textbf{78.3\%} &  \textbf{49.0\%} &  \textbf{42.2\%} & {31.7\%} \\
			\hline
			\multirow{9}{*}{Res-152}
			&DI-PGD~\cite{xie2019improving}  		  &51.8\%   &48.1\% & 40.6\% &  / & 9.7\% & 8.3\% &  6.2\% \\
			&M-PGD~\cite{dong2018boosting}  			   &50.2\%  &44.9\% & 39.4\% & / & 13.9\% & 12.0\% & 7.8\% \\
			&M-DI-PGD  	  &\textbf{76.2\%} &73.3\% & \textbf{69.5\%} &/ & 24.6\% & 21.1\% & 12.7\% \\
			&SI-N-PGD~\cite{lin2020nesterov}      &59.6\%   &50.1\% & 51.3\% & / & 37.9\% & 34.0\% & 20.7\% \\
			&IR-PGD~\cite{wang2021unified}      &42.3\%   &33.8\% & 34.1\% & / & 22.0\% & 20.6\% & 16.2\% \\
			&FIA~\cite{wang2021feature}      &73.8\%   &67.2\% & 67.9\% & / & 48.0\% &43.7\% & 30.4\% \\
			&\textbf{MTA-PGD}   & 70.7\%  & {77.5\%} &  62.8\% & / &  {53.0\%} &  {59.2\%} & {56.3\%} \\
			&\textbf{MTA-IR-PGD}   & 72.8\%  & \textbf{78.0\%} &  64.3\% & / &  \textbf{54.9\%} &  \textbf{63.0\%} & \textbf{59.3\%} \\
			\cline{2-9}
			&SGM-PGD~\cite{wu2020skip}$_{\epsilon=16}^*$      &57.2\%   &48.6\% & 45.4\% & / & 31.6\% & 27.8\% & 20.0\% \\
			&IR-PGD~\cite{wang2021unified}$_{\epsilon=16}^*$      &53.6\%   &50.6\% & 46.0\% & / & / & / & / \\
			&\textbf{MTA-PGD$_{\epsilon=16}$}  	& \textbf{76.0\%}  & \textbf{80.5\%} &  \textbf{67.6\%} &  / &  \textbf{60.5\%} &  \textbf{68.4\%}& \textbf{62.6\%} \\
			\bottomrule
		\end{tabular}
	}
	\label{tab:imagenet_one}
\end{table*}

\vspace{-10pt}
\subsubsection{Experimental results}
\vspace{-2pt}
Table~\ref{tab:cifar} shows the experimental results.
The recently proposed white-box attack method A-PGD~\cite{croce2020reliable} is also treated as a compared method here.
Apparently, MTA-PGD performs much better than all the previous methods with significantly increased transfer attack success rates.
For example, compared with IR-PGD, MTA-PGD improves the success rates by 54.8\%, 26.3\%, 45.4\%, 32.5\%, 35.7\%, 23.3\%, 30.9\%, and 34.7\% on the eight target models.
The results of MTA-PGD$_{\gamma_1=0}$, MTA-PGD$_{\gamma_2=0}$, and MTA-PGD$_{dense}$ will be discussed in ablation study (Section~\ref{sec:ablation}).

\vspace{-5pt}
\subsection{Experiments on Imagenet}
\vspace{-3pt}
\subsubsection{Experimental configurations}
\vspace{-3pt}
We directly use the public trained ImageNet models\footnote{https://github.com/pudae/tensorflow-densenet}$^,$\footnote{https://github.com/tensorflow/models/tree/r1.12.0/research/slim}$^,$\footnote{https://github.com/tensorflow/models/tree/r1.12.0/research/adv\_ \\imagenet\_models} including ResNet-50, -101, -152~\cite{he2016deep}, DenseNet-121, -161~\cite{huang2017densely}, Inception-V3~\cite{szegedy2016rethinking}, -V4~\cite{szegedy2017inception}, Inception-ResNet-V2, Inception-V3$_{ens3}$, Inception-V3$_{ens4}$, and Inception-ResNet-V2$_{ens}$. 
The former eight models are normally trained models while the latter three are secure models trained by ensemble adversarial training~\cite{tramer2017ensemble}.
We shorten these models as Res-50, Res-101, Res-152, DN-121, DN-161, Inc-V3, Inc-V4, IncRes-V2, Inc-V3$_{ens3}$, Inc-V3$_{ens4}$, and IncRes-V3$_{ens}$.

\noindent \textbf{Training the MSM}.
The default network architecture of the MSM is ResNet-19 shown in Figure \ref{fig:network}, with $M1$, $M2$, $M3$, and $M4$ set to 32, 80, 200, and 500, respectively.
We follow previous works \cite{dong2018boosting,wu2020skip} to evaluate the transferability of AEs in two settings: using a single source model and using multiple source models.
We set the input resolution of the MSM to 224$\times$224.
Note that, when the resolution of the source model differs from that of the MSM, we resize the AE $x_{adv}^{T}$ to the resolution of the source model before feeding it into the source model.
More details about training the MSM will be shown in Appendix.

\noindent \textbf{Evaluating the MSM}.
We follow the compared methods 
to randomly choose 5,000 validation images from ImageNet that are correctly classified by all models for evaluation.
Note that, when the resolutions of the MSM and the target model are different, we resize the AE $x_{adv}^{T}$ to the resolution of the target model.
For instance, when attacking Inc-V3 whose resolution is 299$\times$299, we first resize $x_{adv}^{T}$ from 224$\times$224 to 299$\times$299 and then use the resized $x_{adv}^{T}$ to attack Inc-V3.

\begin{table*}[t]
	\centering
	\caption{The transfer attack success rates on seven black-box networks where we use multiple source models.
	}
	{\small 
		\begin{tabular}{ccccccccc}
			\toprule
			Source & Method & Inc-V3 & Inc-V4 & IncRes-V2 & Res-101 & Inc-V3$_{ens3}$ & Inc-V3$_{ens4}$ & IncRes-V2$_{ens}$ \\
			\hline
			\multirow{7}{*}{\tabincell{c}{Res-50 \\ + \\ Res-152 \\ + \\ DN-161}}
			&DI-PGD~\cite{xie2019improving}  		  &86.9\%   &84.3\% & 81.8\% & 96.7\% & 59.7\% & 55.1\% & 41.9\% \\
			&M-PGD~\cite{dong2018boosting}  				&82.0\%  &76.1\% & 76.0\% & 98.0\% & 63.6\% & 60.3\% & 49.6\% \\
			&TI-PGD~\cite{dong2019evading}  				&47.2\%  &44.3\% &37.5\% &85.9\% & 41.7\% & 42.8\% & 33.5\% \\
			&TI-DIM~\cite{dong2019evading}                 &60.7\% & 59.3\% & 50.2\% & 86.8\% & 54.9\% & 56.2\% & 46.9\% \\
			&SGM-PGD~\cite{wu2020skip}  	 &81.8\%   &74.7\% & 73.9\% & \textbf{98.7\%} & 54.9\% & 50.1\% & 38.7\% \\
			&SGM-DI-PGD  	 &86.2\%   &83.9\% & 81.6\% & {98.3\%} & 69.8\% & 64.9\% & 54.4\% \\
			&SGM-M-PGD  	 &86.5\%   &84.3\% & 82.7\% & {98.2\%} & 71.1\% & 67.4\% &  60.8\% \\
			&IR-PGD~\cite{wang2021unified}  	 &75.2\%   &70.3\% & 67.9\% & {90.6\%} & 51.7\% & 49.1\% & 37.5\% \\
			&\textbf{MTA-PGD}  	  &  {90.4\%}  & {94.3\%} &  {87.6\%} &  {97.5\%} &  {75.5\%} &  {79.7\%} & {79.0\%} \\
			&\textbf{MTA-IR-PGD}  	  &  \textbf{93.1\%}  & \textbf{95.8\%} &  \textbf{90.5\%} &  {98.3\%} &  \textbf{83.6\%} &  \textbf{87.2\%} & \textbf{85.0\%} \\
			\hline
			\multirow{7}{*}{\tabincell{c}{Res-50 \\ + \\ Inc-V1 \\ + \\ DN-121}}
			&DI-PGD~\cite{xie2019improving}  		  &84.1\%   &82.3\% & {79.4\%} & {93.9\%} & 56.3\% & 50.1\% & 35.2\% \\
			&M-PGD~\cite{dong2018boosting}  				&79.9\%  &73.6\% & 72.3\% & 93.7\% & 59.3\% & 56.0\% & 42.7\% \\
			&TI-PGD~\cite{dong2019evading}  				&45.0\%  &41.4\% & 33.6\% & 70.2\% & 37.9\% & 38.6\% & 27.1\% \\
			&TI-DIM~\cite{dong2019evading}      & 61.9\% & 58.5\%  & 49.0\%  &79.7\% & 53.1\% & 54.0\% &41.9\% \\
			&SGM-PGD~\cite{wu2020skip}  	 &62.7\%   &53.5\% & 50.9\% & 89.1\% & 33.8\% & 30.4\% & 19.3\% \\
			&SGM-DI-PGD  	 &87.2\%   &83.6\% & \textbf{79.5\%} & {95.1\%} & 59.6\% & 54.9\% & 37.9\% \\
			&SGM-M-PGD  	 &82.8\%   &76.0\% & 74.3\% & \textbf{95.9\%} & 62.2\% & 59.7\% & 45.3\% \\
			&IR-PGD~\cite{wang2021unified}  	 &76.5\%   &70.9\% & 64.0\%  &92.1\% & 51.3\% & 44.9\% & 31.5\% \\
			&\textbf{MTA-PGD}  	  &  {91.7\%}  & {86.4\%} &  {76.0\%} &  {93.6\%} &  {81.7\%} &  \textbf{79.6\%} & \textbf{61.6\%} \\
			&\textbf{MTA-IR-PGD}  	  &  \textbf{92.8\%}  & \textbf{87.9\%} &  {77.2\%} &  {93.8\%} &  \textbf{82.6\%} &  {79.3\%} & {61.5\%} \\
			\hline
			\multirow{7}{*}{\tabincell{c}{Res-50 \\ + \\ Inc-V1}}
			&DI-PGD~\cite{xie2019improving}  		  &76.1\%   &69.3\% & 66.3\% & 90.0\% & 43.5\% & 39.2\% & 25.5\% \\
			&M-PGD~\cite{dong2018boosting}  				&69.5\%  &60.1\% & 59.5\% & 91.5\% & 47.1\% & 44.7\% & 32.5\% \\
			&TI-PGD~\cite{dong2019evading}  				&32.8\%  &28.2\% &22.6\% & 60.4\% & 26.8\% & 27.2\% & 19.0\% \\
			&TI-DIM~\cite{dong2019evading}  				&51.7\%  &46.9\% &38.2\% & 73.3\% & 43.3\% & 44.2\% & 32.6\% \\
			&SGM-PGD~\cite{wu2020skip}  	 &46.1\%   &35.6\% & 33.3\% & 82.0\% & 22.1\% & 19.5\% & 12.3\% \\
			&SGM-DI-PGD  	 &79.2\%   &70.6\% & 68.7\% & {91.9\%} & 47.9\% & 42.0\% & 28.1\% \\
			&SGM-M-PGD  	 &71.9\%   &62.0\% & 61.3\% & {94.3\%} & 49.6\% & 47.2\% & 33.8\% \\
			&IR-PGD~\cite{wang2021unified}  	 &60.2\%   &49.0\% & 46.2\% & 93.0\% & 36.5\% & 30.6\% & 21.0\% \\
			&\textbf{MTA-PGD}  	  &  {84.1\%}  & {88.8\%} &  {78.4\%} &  {93.9\%} &  {60.6\%} &  {61.1\%} & {55.1\%} \\
			&\textbf{MTA-IR-PGD}  	  &  \textbf{87.6\%}  & \textbf{91.8\%} &  \textbf{83.9\%} &  \textbf{95.2\%} &  \textbf{71.5\%} &  \textbf{72.6\%} & \textbf{63.7\%} \\
			\bottomrule
		\end{tabular}
	}
	\label{tab:imagenet_multi}
\end{table*}

\vspace{-8pt}
\subsubsection{Using one source model}
\vspace{-2pt}
The experimental results of using one source model are reported in Table \ref{tab:imagenet_one}.
Note that, in this work, we only focus on the transfer attack testing scene and neglect the white-box attack testing scene.
So we left the results of the testing scenes where the target model is the source model itself to $/$.
M-DI-PGD is a combination of M-PGD and DI-PGD.
IR-PGD is our re-implementation with ${\epsilon=15}$ and the details of the implementation will be shown in Appendix.
Obviously, MTA-PGD outperforms the baselines on almost all testing scenes with great margins, especially when attacking adversarially trained models.
For example, compared with FIA, MTA-PGD improves the transfer attack success rates by about 31.7\%, 30.7\%, 41.1\%, 131.1\%, 41.9\%, and 75.2\% when using the Inc-V3 source model and attacking the target models (Inc-V4, IncRes-152, Res-152, Inc-V3$_{ens3}$, Inc-V3$_{ens4}$, IncRes-V2$_{ens}$).
MTA-IR-PGD is the combination of MTA-PGD and IR-PGD.
Compared with MTA-PGD, MTA-IR-PGD improves the attack success rates by about 5.1\%, 6.8\%, 14.7\%,  23.3\%, 44.8\%, and 55.9\% when using the Inc-V3 source model and attacking the target models, indicating that existing transferable attack methods can further improve MTA-PGD.

Recall that SGM-PGD only works for source models with lots of skip connections (e.g., ResNet). And the original paper sets $\epsilon$ to 16, which differs from most of the other methods.
The official IR-PGD also sets $\epsilon$ to 16.
Therefore, we copy their results with $\epsilon=16$ from their official paper to Table \ref{tab:imagenet_one} and denote them as SGM-PGD$_{\epsilon=16}^*$ and IR-PGD$_{\epsilon=16}^*$, respectively.
To compare MTA-PGD with them, we further set $\epsilon$ to 16 for MTA-PGD and denote the new result as MTA-PGD$_{\epsilon=16}$.
The comparisons show that MTA-PGD$_{\epsilon=16}$ outperforms SGM-PGD$_{\epsilon=16}^*$ and IR-PGD$_{\epsilon=16}^*$ significantly.

When using IncRes-V2 source model, MTA-PGD sometimes performs slightly worse than M-DI-PGD, possibly because the MSM with ResNet-19 backbone is not suitable to be trained to attack IncRes-V2.
We then replace the backbone from ResNet-19 with another simplified Inception network (the architecture will be shown in Appendix) and retrain the MSM. 
The newly trained MSM is denoted as MTA-PGD$_{Inc}$.
Compared with ResNet-19, the simplified Inception backbone is more similar to IncRes-V2 so that MTA-PGD$_{Inc}$ turns to be easier to generate adversarial attacks to fool IncRes-V2 than MTA-PGD, leading to easier convergence of MTA-PGD$_{Inc}$.
The experimental results show that MTA-PGD$_{Inc}$ outperforms not only MTA-PGD but also the compared methods in most testing scenes, indicating 1) the advantage of the proposed MTA framework and 2) MTA-PGD can be further improved by using more suitable backbones.

\vspace{-10pt}
\subsubsection{Using multiple source models}
\vspace{-2pt}
The experimental results of using multiple source models are reported in Table \ref{tab:imagenet_multi}.
We use three source model groups (Res-50+Res-152+DN161, Res-50+Inc-V1+DN-121, Res-50+Inc-V1) to train the MSM, respectively, and use seven target models (Inc-V3, Inc-V4, InvRes-V2, Res-101, Inc-V3$_{ens3}$, Inc-V3$_{ens4}$, IncRes-V2$_{ens}$) to evaluate the transferability of the attacks to the MSM.
SGM-X-PGD is the combination of SGM-PGD and X-PGD (X=DI or M).
MTA-IR-PGD is the combination of MTA-PGD and IR-PGD.
The results show that MTA-PGD outperforms the baselines in almost all testing scenes, especially when attacking defensive models.
For instance, compared with SGM-DI-PGD, MTA-PGD improves the transfer attack success rates by 6.8\%, 25.8\%, 14.1\%, 2.2\%, 26.5\%, 45.5\%, and 96.1\% on the seven target models when using Res-50 and Inc-V1 source models. 
Besides, MTA-IR-PGD outperforms MTA-PGD, indicating that existing transferable attack methods can further improve MTA-PGD.

\vspace{-5pt}
\subsection{Ablation Study}
\label{sec:ablation}
\vspace{-2pt}
\subsubsection{Network structure}
\vspace{-2pt}
\label{sec:ablation_network}
The comparison between MTA-PGD and MTA-PGD$_{Inc}$ shown in Table~\ref{tab:imagenet_one} has validated the effect of backbone on the MSM.
Here we conduct another experiment on Cifar-10 to further verify the effect of backbone by replacing the backbone from ResNet-13 to DenseNet-22BC (the structure of DenseNet-22BC will be shown in Appendix).
We denote the MSM using DenseNet-22BC backbone as MTA-PGD$_{dense}$ and report its experimental results in Table \ref{tab:cifar}. 
The comparisons among MTA-PGD, MTA-PGD$_{dense}$, and the other compared methods indicate that 1) the backbone affects the performance of MTA-PGD; 2) MTA-PGD outperforms the compared methods with various backbones.
It also may inspire us to design more suitable backbones to improve MTA-PGD.

\vspace{-5pt}
\subsubsection{Number of attack iterations}
\vspace{-3pt}
\label{sec:ablation_T}

\begin{figure}[]
	\centering
	\includegraphics[width=0.469\textwidth]{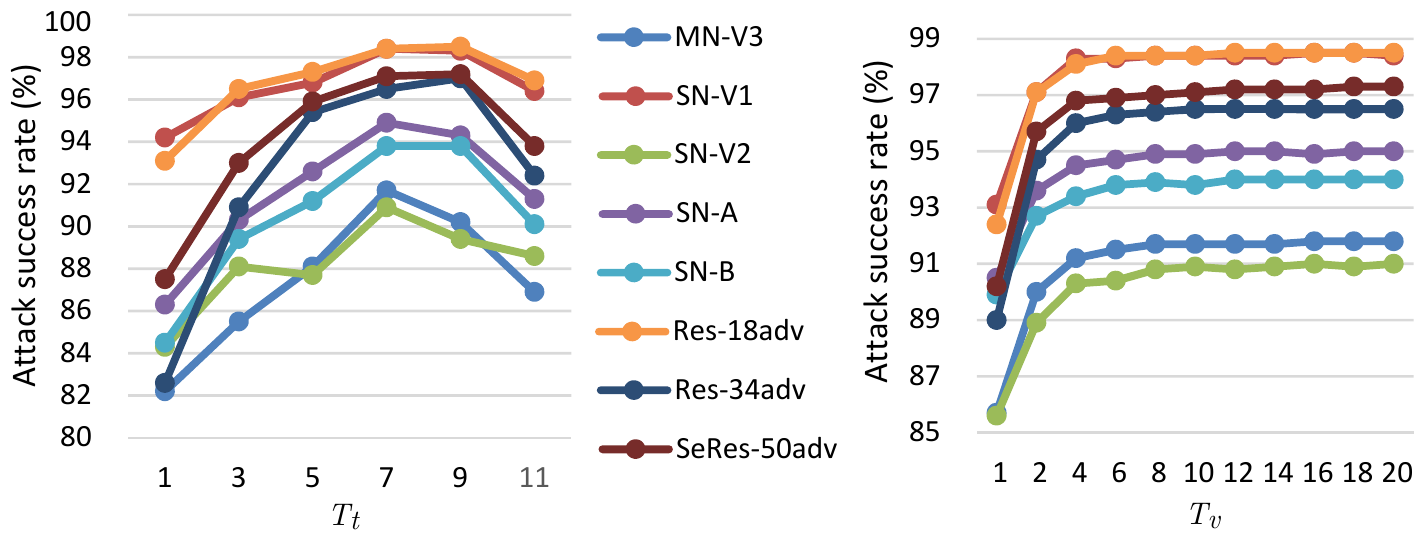}
	\vspace{-1pt}
	\caption{
		Transfer attack performances of MTA-PGD on the eight Cifar-10 target models.
		Left: Attack success rates with different $T_t$. 
		Right: Attack success rates with different $T_v$. 
	}
	\vspace{-10pt}
	\label{fig:ablation-T}
\end{figure}

We perform several experiments on Cifar-10 to validate how the number of attack iterations $T_t$ affects the performance.
$T_t$ is set to 7 by default on Cifar-10.
Here we set $T_t$ to 1, 3, 5, 9, and 11 and keep all the other settings be consistent with the default settings.
Figure \ref{fig:ablation-T} shows the corresponding performances of MTA-PGD.
It is observed that when $T_t<7$, the performances of MTA-PGD will be improved with the increase of $T_t$ while when $T_t>7$, the performance tends to drop. 
We think this is due to the difficulty of unrolling too many attack steps when training the MSM.

We also verify how $T_v$ affects the performance by changing $T_v$.
$T_v$ is default set to 10 in all our experiments.
Figure \ref{fig:ablation-T} shows the experimental results using different numbers of $T_v$.
When $T_v=1$, the performances can be denoted as MTA-FGSM (one-step PGD).
With the increase of $T_v$, the transfer attack success rates increase clearly.

\vspace{-5pt}
\subsubsection{The effects of $\gamma_1$ and $\gamma_2$}
\vspace{-2pt}
\label{sec:ablation_gamma}
We perform two experiments on Cifar-10 to verify how the parameters $\gamma_1$ and $\gamma_2$ in Eq~\eqref{eq:my_PGD} affect the transfer attack performance.
In the two experiments, we set $\gamma_1$ and $\gamma_2$ to zero respectively, and amplify $\epsilon_c$ appropriately to offset the decrease of the training perturbation size caused by zeroing $\gamma_1$ or $\gamma_2$.
We denote the two newly performed MTA-PGD as MTA-PGD$_{\gamma_1=0}$ and MTA-PGD$_{\gamma_2=0}$. 
Table \ref{tab:cifar} shows the experimental results. 
The results show that by setting $\gamma_1$ to zero, the performances of MTA-PGD are greatly damaged by more than 17\% on all target models, indicating the indispensability of the arctan component in the Customized PGD.
Setting $\gamma_2$ to zero also decreases MTA-PGD's performances, but the effect is smaller than that of $\gamma_1$.
Overall, the two experiments demonstrate the indispensability of Customized PGD for the proposed MTA framework.
Further, both the arctan and sign components in Customized PGD are important to train the MSM, especially arctan.

\vspace{-7pt}
\section{Conclusion}
\vspace{-5pt}
Existing query free black-box adversarial attack methods directly use image classification models as surrogate models to generate transferable adversarial attacks to attack black-box models neglecting the study of surrogate models.
In this paper, we propose a novel framework called meta-transfer attack (MTA) to improve the transferability of adversarial attacks via training an MSM using these surrogate models.
The MSM is a particular model trained to learn how to make the adversarial attacks to it can fool the surrogate models.
To enable and improve the training of the MSM, a novel Customized PGD is also developed.
Through extensive experiments, we validate that by attacking the trained MSM, we can get transferable adversarial attacks that are generalizable to attack black-box target models with much higher success rates than existing methods, demonstrating the effectiveness of the proposed MTA framework.
Our work is promising to evaluate and improve the security of deep models, and has no potential negative societal impacts.


{\small
	\bibliographystyle{ieee_fullname}
	\bibliography{main}

\begin{thebibliography}{10}\itemsep=-1pt

\bibitem{abadi2016tensorflow}
Mart{\'\i}n Abadi, Paul Barham, Jianmin Chen, Zhifeng Chen, Andy Davis, Jeffrey
  Dean, Matthieu Devin, Sanjay Ghemawat, Geoffrey Irving, Michael Isard, et~al.
\newblock Tensorflow: A system for large-scale machine learning.
\newblock In {\em 12th $\{$USENIX$\}$ symposium on operating systems design and
  implementation ($\{$OSDI$\}$ 16)}, pages 265--283, 2016.

\bibitem{bose2020adversarial}
Avishek~Joey Bose, Gauthier Gidel, Hugo Berrard, Andre Cianflone, Pascal
  Vincent, Simon Lacoste-Julien, and William~L Hamilton.
\newblock Adversarial example games.
\newblock {\em Advances in neural information processing systems}, 2020.

\bibitem{brendel2017decision}
Wieland Brendel, Jonas Rauber, and Matthias Bethge.
\newblock Decision-based adversarial attacks: Reliable attacks against
  black-box machine learning models.
\newblock {\em arXiv preprint arXiv:1712.04248}, 2017.

\bibitem{carlini2017towards}
Nicholas Carlini and David Wagner.
\newblock Towards evaluating the robustness of neural networks.
\newblock In {\em 2017 ieee symposium on security and privacy (sp)}, pages
  39--57. IEEE, 2017.

\bibitem{chen2020hop}
Jianbo Chen, Michael~I Jordan, and Martin~J Wainwright.
\newblock Hop{S}kip{J}ump{A}ttack: a query-efficient decision-based adversarial
  attack.
\newblock In {\em 2020 IEEE Symposium on Security and Privacy (SP)}. IEEE,
  2020.

\bibitem{chen2017zoo}
Pin-Yu Chen, Huan Zhang, Yash Sharma, Jinfeng Yi, and Cho-Jui Hsieh.
\newblock Zoo: Zeroth order optimization based black-box attacks to deep neural
  networks without training substitute models.
\newblock In {\em Proceedings of the 10th ACM workshop on artificial
  intelligence and security}, pages 15--26, 2017.

\bibitem{cheng2018query}
Minhao Cheng, Thong Le, Pin-Yu Chen, Jinfeng Yi, Huan Zhang, and Cho-Jui Hsieh.
\newblock Query-efficient hard-label black-box attack: An optimization-based
  approach.
\newblock {\em arXiv preprint arXiv:1807.04457}, 2018.

\bibitem{cheng2020signopt}
Minhao Cheng, Simranjit Singh, Patrick~H. Chen, Pin-Yu Chen, Sijia Liu, and
  Cho-Jui Hsieh.
\newblock Sign-opt: A query-efficient hard-label adversarial attack.
\newblock In {\em international conference on learning representations}, 2020.

\bibitem{croce2020minimally}
Francesco Croce and Matthias Hein.
\newblock Minimally distorted adversarial examples with a fast adaptive
  boundary attack.
\newblock In {\em International Conference on Machine Learning}, pages
  2196--2205. PMLR, 2020.

\bibitem{croce2020reliable}
Francesco Croce and Matthias Hein.
\newblock Reliable evaluation of adversarial robustness with an ensemble of
  diverse parameter-free attacks.
\newblock In {\em International Conference on Machine Learning}, pages
  2206--2216. PMLR, 2020.

\bibitem{deng2009imagenet}
Jia Deng, Wei Dong, Richard Socher, Li-Jia Li, Kai Li, and Li Fei-Fei.
\newblock Imagenet: A large-scale hierarchical image database.
\newblock In {\em 2009 IEEE conference on computer vision and pattern
  recognition}, pages 248--255. Ieee, 2009.

\bibitem{dong2018boosting}
Yinpeng Dong, Fangzhou Liao, Tianyu Pang, Hang Su, Jun Zhu, Xiaolin Hu, and
  Jianguo Li.
\newblock Boosting adversarial attacks with momentum.
\newblock In {\em Proceedings of the IEEE conference on computer vision and
  pattern recognition}, pages 9185--9193, 2018.

\bibitem{dong2019evading}
Yinpeng Dong, Tianyu Pang, Hang Su, and Jun Zhu.
\newblock Evading defenses to transferable adversarial examples by
  translation-invariant attacks.
\newblock In {\em Proceedings of the IEEE/CVF Conference on Computer Vision and
  Pattern Recognition}, pages 4312--4321, 2019.

\bibitem{du2020query-efficient}
Jiawei Du, Hu Zhang, Tianyi~Joey Zhou, Yi Yang, and Jiashi Feng.
\newblock Query-efficient meta attack to deep neural networks.
\newblock {\em International Conference on Learning Representations}, 2020.

\bibitem{finn2017model}
Chelsea Finn, Pieter Abbeel, and Sergey Levine.
\newblock Model-agnostic meta-learning for fast adaptation of deep networks.
\newblock In {\em International Conference on Machine Learning}, pages
  1126--1135. PMLR, 2017.

\bibitem{ganeshan2019fda}
Aditya Ganeshan, Vivek BS, and R~Venkatesh Babu.
\newblock Fda: Feature disruptive attack.
\newblock In {\em Proceedings of the IEEE/CVF International Conference on
  Computer Vision}, pages 8069--8079, 2019.

\bibitem{gao2020patch}
Lianli Gao, Qilong Zhang, Jingkuan Song, Xianglong Liu, and Heng~Tao Shen.
\newblock Patch-wise attack for fooling deep neural network.
\newblock In {\em European Conference on Computer Vision}, pages 307--322.
  Springer, 2020.

\bibitem{goodfellow2014explaining}
Ian~J Goodfellow, Jonathon Shlens, and Christian Szegedy.
\newblock Explaining and harnessing adversarial examples.
\newblock {\em international conference on learning representations}, 2014.

\bibitem{he2016deep}
Kaiming He, Xiangyu Zhang, Shaoqing Ren, and Jian Sun.
\newblock Deep residual learning for image recognition.
\newblock In {\em Proceedings of the IEEE conference on computer vision and
  pattern recognition}, pages 770--778, 2016.

\bibitem{howard2019searching}
Andrew Howard, Mark Sandler, Grace Chu, Liang-Chieh Chen, Bo Chen, Mingxing
  Tan, Weijun Wang, Yukun Zhu, Ruoming Pang, Vijay Vasudevan, et~al.
\newblock Searching for mobilenetv3.
\newblock In {\em Proceedings of the IEEE/CVF International Conference on
  Computer Vision}, pages 1314--1324, 2019.

\bibitem{howard2017mobilenets}
Andrew~G Howard, Menglong Zhu, Bo Chen, Dmitry Kalenichenko, Weijun Wang,
  Tobias Weyand, Marco Andreetto, and Hartwig Adam.
\newblock Mobilenets: Efficient convolutional neural networks for mobile vision
  applications.
\newblock {\em arXiv preprint arXiv:1704.04861}, 2017.

\bibitem{hu2018squeeze}
Jie Hu, Li Shen, and Gang Sun.
\newblock Squeeze-and-excitation networks.
\newblock In {\em Proceedings of the IEEE conference on computer vision and
  pattern recognition}, pages 7132--7141, 2018.

\bibitem{huang2017densely}
Gao Huang, Zhuang Liu, Laurens Van Der~Maaten, and Kilian~Q Weinberger.
\newblock Densely connected convolutional networks.
\newblock In {\em Proceedings of the IEEE conference on computer vision and
  pattern recognition}, pages 4700--4708, 2017.

\bibitem{huang2019enhancing}
Qian Huang, Isay Katsman, Horace He, Zeqi Gu, Serge Belongie, and Ser-Nam Lim.
\newblock Enhancing adversarial example transferability with an intermediate
  level attack.
\newblock In {\em Proceedings of the IEEE/CVF International Conference on
  Computer Vision}, pages 4733--4742, 2019.

\bibitem{huang2020black}
Zhichao Huang and Tong Zhang.
\newblock Black-box adversarial attack with transferable model-based embedding.
\newblock {\em International Conference on Learning Representations}, 2020.

\bibitem{iandola2016squeezenet}
Forrest~N Iandola, Song Han, Matthew~W Moskewicz, Khalid Ashraf, William~J
  Dally, and Kurt Keutzer.
\newblock Squeezenet: Alexnet-level accuracy with 50x fewer parameters and< 0.5
  mb model size.
\newblock {\em arXiv preprint arXiv:1602.07360}, 2016.

\bibitem{ilyas2018black}
Andrew Ilyas, Logan Engstrom, Anish Athalye, and Jessy Lin.
\newblock Black-box adversarial attacks with limited queries and information.
\newblock In {\em International Conference on Machine Learning}, pages
  2137--2146. PMLR, 2018.

\bibitem{ilyas2019adversarial}
Andrew Ilyas, Shibani Santurkar, Dimitris Tsipras, Logan Engstrom, Brandon
  Tran, and Aleksander Madry.
\newblock Adversarial examples are not bugs, they are features.
\newblock {\em arXiv preprint arXiv:1905.02175}, 2019.

\bibitem{krizhevsky2009learning}
Alex Krizhevsky, Geoffrey Hinton, et~al.
\newblock Learning multiple layers of features from tiny images.
\newblock 2009.

\bibitem{krizhevsky2012imagenet}
Alex Krizhevsky, Ilya Sutskever, and Geoffrey~E Hinton.
\newblock Imagenet classification with deep convolutional neural networks.
\newblock {\em Advances in neural information processing systems},
  25:1097--1105, 2012.

\bibitem{kurakin2016adversarial}
Alexey Kurakin, Ian Goodfellow, Samy Bengio, et~al.
\newblock Adversarial examples in the physical world, 2016.

\bibitem{lecun1995convolutional}
Yann LeCun, Yoshua Bengio, et~al.
\newblock Convolutional networks for images, speech, and time series.
\newblock {\em The handbook of brain theory and neural networks},
  3361(10):1995, 1995.

\bibitem{li2020qeba}
Huichen Li, Xiaojun Xu, Xiaolu Zhang, Shuang Yang, and Bo Li.
\newblock Qeba: Query-efficient boundary-based blackbox attack.
\newblock In {\em Proceedings of the IEEE/CVF Conference on Computer Vision and
  Pattern Recognition}, pages 1221--1230, 2020.

\bibitem{lin2020nesterov}
Jiadong Lin, Chuanbiao Song, Kun He, Liwei Wang, and John~E Hopcroft.
\newblock Nesterov accelerated gradient and scale invariance for adversarial
  attacks.
\newblock {\em International Conference on Learning Representations}, 2020.

\bibitem{liu2016delving}
Yanpei Liu, Xinyun Chen, Chang Liu, and Dawn Song.
\newblock Delving into transferable adversarial examples and black-box attacks.
\newblock {\em international conference on learning representations}, 2017.

\bibitem{madry2017towards}
Aleksander Madry, Aleksandar Makelov, Ludwig Schmidt, Dimitris Tsipras, and
  Adrian Vladu.
\newblock Towards deep learning models resistant to adversarial attacks.
\newblock {\em international conference on learning representations}, 2018.

\bibitem{moosavi2016deepfool}
Seyed-Mohsen Moosavi-Dezfooli, Alhussein Fawzi, and Pascal Frossard.
\newblock Deepfool: a simple and accurate method to fool deep neural networks.
\newblock In {\em Proceedings of the IEEE conference on computer vision and
  pattern recognition}, pages 2574--2582, 2016.

\bibitem{papernot2016transferability}
Nicolas Papernot, Patrick McDaniel, and Ian Goodfellow.
\newblock Transferability in machine learning: from phenomena to black-box
  attacks using adversarial samples.
\newblock {\em arXiv preprint arXiv:1605.07277}, 2016.

\bibitem{papernot2017practical}
Nicolas Papernot, Patrick McDaniel, Ian Goodfellow, Somesh Jha, Z~Berkay Celik,
  and Ananthram Swami.
\newblock Practical black-box attacks against machine learning.
\newblock In {\em Proceedings of the 2017 ACM on Asia conference on computer
  and communications security}, pages 506--519, 2017.

\bibitem{paszke2017automatic}
Adam Paszke, Sam Gross, Soumith Chintala, Gregory Chanan, Edward Yang, Zachary
  DeVito, Zeming Lin, Alban Desmaison, Luca Antiga, and Adam Lerer.
\newblock Automatic differentiation in pytorch.
\newblock 2017.

\bibitem{qin2020layer}
Yunxiao Qin, Weiguo Zhang, Zezheng Wang, Chenxu Zhao, and Jingping Shi.
\newblock Layer-wise adaptive updating for few-shot image classification.
\newblock {\em IEEE Signal Processing Letters}, 27:2044--2048, 2020.

\bibitem{ren2015faster}
Shaoqing Ren, Kaiming He, Ross Girshick, and Jian Sun.
\newblock Faster r-cnn: towards real-time object detection with region proposal
  networks.
\newblock {\em IEEE transactions on pattern analysis and machine intelligence},
  39(6):1137--1149, 2016.

\bibitem{sandler2018mobilenetv2}
Mark Sandler, Andrew Howard, Menglong Zhu, Andrey Zhmoginov, and Liang-Chieh
  Chen.
\newblock Mobilenetv2: Inverted residuals and linear bottlenecks.
\newblock In {\em Proceedings of the IEEE conference on computer vision and
  pattern recognition}, pages 4510--4520, 2018.

\bibitem{szegedy2017inception}
Christian Szegedy, Sergey Ioffe, Vincent Vanhoucke, and Alexander Alemi.
\newblock Inception-v4, inception-resnet and the impact of residual connections
  on learning.
\newblock In {\em Proceedings of the AAAI Conference on Artificial
  Intelligence}, volume~31, 2017.

\bibitem{szegedy2016rethinking}
Christian Szegedy, Vincent Vanhoucke, Sergey Ioffe, Jon Shlens, and Zbigniew
  Wojna.
\newblock Rethinking the inception architecture for computer vision.
\newblock In {\em Proceedings of the IEEE conference on computer vision and
  pattern recognition}, pages 2818--2826, 2016.

\bibitem{szegedy2014intriguing}
Christian Szegedy, Wojciech Zaremba, Ilya Sutskever, Joan Bruna, Dumitru Erhan,
  J.~Ian Goodfellow, and Rob Fergus.
\newblock Intriguing properties of neural networks.
\newblock {\em international conference on learning representations}, 2014.

\bibitem{tramer2017ensemble}
Florian Tram{\`e}r, Alexey Kurakin, Nicolas Papernot, Ian Goodfellow, Dan
  Boneh, and Patrick McDaniel.
\newblock Ensemble adversarial training: Attacks and defenses.
\newblock {\em arXiv preprint arXiv:1705.07204}, 2017.

\bibitem{wang2020spanning}
Lu Wang, Huan Zhang, Jinfeng Yi, Cho-Jui Hsieh, and Yuan Jiang.
\newblock Spanning attack: reinforce black-box attacks with unlabeled data.
\newblock {\em Machine Learning}, 109(12):2349--2368, 2020.

\bibitem{wang2021unified}
Xin Wang, Jie Ren, Shuyun Lin, Xiangming Zhu, Yisen Wang, and Quanshi Zhang.
\newblock A unified approach to interpreting and boosting adversarial
  transferability.
\newblock {\em International Conference on Learning Representations}, 2021.

\bibitem{wang2021feature}
Zhibo Wang, Hengchang Guo, Zhifei Zhang, Wenxin Liu, Zhan Qin, and Kui Ren.
\newblock Feature importance-aware transferable adversarial attacks.
\newblock In {\em Proceedings of the IEEE/CVF International Conference on
  Computer Vision}, 2021.

\bibitem{wu2020skip}
Dongxian Wu, Yisen Wang, Shu-Tao Xia, James Bailey, and Xingjun Ma.
\newblock Skip connections matter: On the transferability of adversarial
  examples generated with resnets.
\newblock {\em international conference on learning representations}, 2020.

\bibitem{wu2018understanding}
Lei Wu, Zhanxing Zhu, Cheng Tai, et~al.
\newblock Understanding and enhancing the transferability of adversarial
  examples.
\newblock {\em arXiv preprint arXiv:1802.09707}, 2018.

\bibitem{xie2019improving}
Cihang Xie, Zhishuai Zhang, Yuyin Zhou, Song Bai, Jianyu Wang, Zhou Ren, and
  Alan~L Yuille.
\newblock Improving transferability of adversarial examples with input
  diversity.
\newblock In {\em Proceedings of the IEEE/CVF Conference on Computer Vision and
  Pattern Recognition}, pages 2730--2739, 2019.

\bibitem{yuan2021meta}
Zheng Yuan, Jie Zhang, Yunpei Jia, Chuanqi Tan, Tao Xue, and Shiguang Shan.
\newblock Meta gradient adversarial attack.
\newblock In {\em Proceedings of the IEEE/CVF International Conference on
  Computer Vision}, 2021.

\bibitem{zhang2018shufflenet}
Xiangyu Zhang, Xinyu Zhou, Mengxiao Lin, and Jian Sun.
\newblock Shufflenet: An extremely efficient convolutional neural network for
  mobile devices.
\newblock {\em computer vision and pattern recognition}, 2018.

\bibitem{zhou2018transferable}
Wen Zhou, Xin Hou, Yongjun Chen, Mengyun Tang, Xiangqi Huang, Xiang Gan, and
  Yong Yang.
\newblock Transferable adversarial perturbations.
\newblock In {\em Proceedings of the European Conference on Computer Vision
  (ECCV)}, pages 452--467, 2018.

\end{thebibliography}
}

\appendix
\section{Appendix}
\subsection{Testing pseudo code of MTA}
\vspace{-2pt}
We summarize the testing pseudo code of MTA in Algorithm.\ref{algorithm:test}, where $\mathcal{\hat{F}}$ is the target model and $\tilde{y}$ is the target model's prediction for the adversarial example $x_{adv}^{T}$.
Note that all the clean examples in $\mathbb{\hat{D}}$ are correctly classified by the target model.
Len($\mathbb{\hat{D}}$) denotes the number of examples in $\mathbb{\hat{D}}$.

\begin{algorithm}[H]
	\caption{Testing of Meta-Transfer Attack}
	{\bfseries input:} Black-box target model $\mathcal{\hat{F}}$, Testing examples $\mathbb{\hat{D}}$ that are correctly classified by the target model, Optimized meta-surrogate model $\mathcal{M}_\theta$. \\
	{\bfseries output:} Transfer attack success rate. \\
	{\bfseries 1 \;\!:} $P=0$ \\
	{\bfseries 2 \;\!:} \textbf{for} ($x, y$) $\in \mathbb{\hat{D}}$ \ \textbf{do}\\
	{\bfseries 3 \;\!:} \quad $x_{adv}^{0} = x$ \\
	{\bfseries 4 \;\!:} \quad \textbf{for} k in [1, 2, ..., T] \ \textbf{do} \\
	{\bfseries 5 \;\!:} \quad    \quad $g^k = \nabla_{x_{adv}^{k-1}} \emph{L}(\mathcal{M}_\theta(x_{adv}^{k-1}), y)$ \\
	{\bfseries 6 \;\!:} \quad    \quad $x_{adv}^{k} \!=\! \text{Clip}\big(x_{adv}^{k-1} \!+\! \frac{\epsilon}{T} \cdot \text{sign}(g^k)\big)$ \\
	{\bfseries 7 \;\!:} \quad    \textbf{end for} \\
	{\bfseries 8 \;\!:} \quad	evaluate $x_{adv}^{T}$ on $\mathcal{\hat{F}}$ and obtain $\tilde{y} = \mathcal{\hat{F}}(x_{adv}^{T})$   \\
	{\bfseries 9 \;\!:} 	\quad  \textbf{if} $y \neq \tilde{y}$ \ \textbf{do}\\
	{\bfseries 10:} 	\quad \quad $P+=1$ \\
	{\bfseries 11:}  \quad \textbf{end if}\\
	{\bfseries 12:} 	{\bfseries return} $\frac{P}{\text{Len}(\mathbb{\hat{D}})}$
	\label{algorithm:test}
\end{algorithm}

\vspace{-2pt}
\subsection{Training the source and target models on Cifar-10}
\vspace{-2pt}
On Cifar-10, we use 16 source and target models to train and test the meta-surrogate model (MSM).
The 8 source models are ResNet-10, -18, -34, SeResNet-14, -26, -50, MobileNet-V1, and -V2.
The 8 target models are MobileNet-V3, ShuffleNet-V1, -V2, SqueezeNet-A, -B, and adversarially trained ResNet-18, -34 and SeResNet-50.
It is not easy to collect the 16 trained Cifar-10 models on the internet.
Therefore, before the experiments of MTA, we first use consistent hyper-parameters to train the 16 models on Cifar-10 for 200 epochs.
The learning rate, L2 weight decay, and batch size are set to 0.01, 1e-5, and 128, respectively.
For each adversarially trained model, we first use FGSM and the normally trained model to generate one adversarial example for each training image with $\epsilon=3$, and then train the model on both clean and adversarial images.
The 8 source models obtain 90.0\%, 91.8\%, 92.6\%, 85.6\%, 88.3\%, 90.5\%, 82.0\%, and 81.8\% accuracies on the test set, and the 8 target models obtain 80.0\%, 82.5\%, 76.4\%, 86.4\%, 86.9\%, 88.9\%, 90.5\%, and 87.5\% accuracies.

\subsection{More experiments on Cifar-10}

\subsubsection{Targeted transfer attack}
We conduct targeted transfer attack and show the experimental results in Table \ref{tab:targeted attack}. MTA-PGD has a great advantage over the compared methods in the targeted transfer attack setting.

\begin{table}[]
	\centering
	\caption{Targeted transfer attack results on Cifar-10.}
	{\small 
		\begin{tabular}{cccccc}
			\toprule
			Method  & MN-V3 & SN-V1 & SN-V2 & SN-A & SN-B\\
			\midrule
			DI-PGD  		 & 16.3\% & 26.4\% & 17.2\% &  22.3\% & 21.6\% \\
			M-PGD  			& 29.6\% & 43.6\% & 29.8\% & 37.1\% & 35.4\% \\
			TI-PGD  			& 17.6\% & 21.1\% & 16.5\% & 26.1\% & 25.8\% \\
			IR-PGD  			& 10.8\% & 19.6\% & 9.5\% & 13.7\% & 12.5\% \\
			AEG  			& 47.2\% & 53.8\% & 36.5\% & 42.6\% & 41.0\% \\
			\textbf{MTA-PGD}  	  &  \textbf{49.0\%} &  \textbf{70.3\%} & \textbf{47.7\%}  & \textbf{60.3\%} & \textbf{58.5\%} \\
			\bottomrule
		\end{tabular}
	}
	\label{tab:targeted attack}
	\vspace{-5pt}
\end{table}

\subsubsection{Transfer attack with smaller $\epsilon$}
We set $\epsilon$ to 8 to evaluate how does MTA-PGD perform with smaller $\epsilon$.
The results shown in Table \ref{tab:8} indicate that MTA-PGD outperforms the compared methods no matter the value of $\epsilon$.

\begin{table}[]
	\centering
	\caption{Transfer attack results with $\epsilon = 8$ on Cifar-10.}
	{\small 
		\begin{tabular}{cccccc}
			\toprule
			Method  & MN-V3 & SN-V1 & SN-V2 & SN-A & SN-B  \\
			\midrule
			DI-PGD  		 & 31.5\% & 42.1\% & 30.0\% & 38.2\% & 36.9\%  \\
			M-PGD  			& 44.2\% & 59.8\% & 43.2\% & 55.7\% & 54.9\%  \\
			TI-PGD  			& 29.5\% & 31.3\% & 29.6\% & 37.7\% & 36.8\%  \\
			IR-PGD  			& 29.2\% & 51.1\% & 35.3\% & 38.5\% & 37.4\%  \\
			AEG  			& 58.0\% & 66.5\% & 50.4\% & 61.9\% & 59.6\%  \\
			\textbf{MTA-PGD}  	  &  \textbf{62.5\%} &  \textbf{79.6\%} & \textbf{58.2\%} &  \textbf{70.5\%} & \textbf{69.3\%} \\
			\bottomrule
		\end{tabular}
	}
	\label{tab:8}
\end{table}

\subsubsection{Comparison between MTA and MetaAttack}
MetaAttack[14] is developed for query-based black-box adversarial attack but not for transfer attack.
We implement MetaAttack in the transfer attack scene on Cifar-10 and compare it with MTA-PGD in Table~\ref{tab:cifar_ablation}.
The comparison indicates that MTA-PGD greatly outperforms MetaAttack in transfer attack.

\subsubsection{More experiments about the Customized PGD}
As introduced in Section \ref{sec:method}, the sign function in the vanilla PGD with L$_{\infty}$ constraint introduces a discrete operation.
This results in that the gradient back-propagating through $\text{sign}$ be zero and further prohibits the training of the MSM. 
We propose the Customized PGD to enable the training of the MSM.
Here we conduct other four experiments to validate the indispensability and the effect of the Customized PGD on the proposed MTA framework.

As PGD with L2 constraint contains no sign, in the first experiment, we use PGD with L2 constraint ($PGD_{L2}$) instead of the Customized PGD to attack the MSM in the training phase and denote the trained MSM as MTA-PGD$_{PGD_{L2}}$.

PGD with L1 constraint also contains no sign.
In the second experiment, we use PGD with L1 constraint ($PGD_{L1}$) to attack the MSM in the training phase and denote the trained MSM as MTA-PGD$_{PGD_{L1}}$.

Both $\gamma_1$ and $\gamma_2$ of the Customized PGD are set to 0.01 by default.
In the third experiment, we set $\gamma_1$ to 0.05.
Note that we decrease $\epsilon_c$ appropriately to offset the increase of the training perturbation size caused by setting $\gamma_1$ to 0.05.
All the other experimental settings are consistent with the default settings.
We denote the MSM trained in this experiment as MTA-PGD$_{\gamma_1=0.05}$.

In the fourth experiment, we set $\gamma_2$ to 0.05 and denote the trained MSM as MTA-PGD$_{\gamma_2=0.05}$.

Table \ref{tab:cifar_ablation} reports all the four experimental results.
We can get three conclusions.
First, directly using $PGD_{L1}$ or $PGD_{L2}$ in MTA's training stage is also effective to train the MSM but leads to limited performance.
Second, the proposed Customized PGD is important for the proposed MTA framework to achieve superior performance.
Third, larger $\gamma_1$ or $\gamma_2$ damages the performances of MTA-PGD.

\begin{table*}[t]
	\centering
	\caption{More transfer attack experimental results on Cifar-10. 
	}
	{  
		\begin{tabular}{cccccc}
			\toprule
			Method  & MN-V3 & SN-V1 & SN-V2 & SN-A & SN-B  \\
			\midrule
			MetaAttack  	  &  39.2\% &  {43.9\%} & {32.1\%} & {38.6\%} & {37.8\%} \\
			\textbf{MTA-PGD$_{PGD_{L2}}$ }  	  &  {80.8\%} &  {92.7\%} & {83.5\%} & {89.0\%} & 86.8\%  \\
			\textbf{MTA-PGD$_{PGD_{L1}}$ }  	  &  {81.5\%} &  {91.3\%} & {82.4\%} & {85.3\%} & 83.7\%  \\
			\textbf{MTA-PGD$_{\gamma_1=0.05}$ }  	  &  90.5\% &  {98.0\%} & {90.2\%} & {94.5\%} & 93.1\%  \\
			\textbf{MTA-PGD$_{\gamma_2=0.05}$ }  	  &  {86.7\%} &  {95.3\%} & {85.8\%} & {89.5\%} & 88.4\%  \\
			\textbf{MTA-PGD }  	  &  \textbf{91.8\%} &  \textbf{98.4\%} & \textbf{90.9\%} & \textbf{94.9\%} & \textbf{93.8\%} \\
			\bottomrule
		\end{tabular}
	}
	\label{tab:cifar_ablation}
\end{table*}

\vspace{-2pt}
\subsection{The supplemental experimental settings of MTA on ImageNet.}
\vspace{-2pt}

In our experiment on ImageNet, we found that the MSM directly trained on the resolution of $224\times 224$ often suffers from slow and unstable convergence due to the high dimensionality. 
Therefore, we develop a three-stage training strategy for gradually and stably training the MSM.
The \textbf{first} training stage only trains the top 4 blocks (blocks 3-6 in Fig.\ref{fig:network}) and the classifier of the MSM.
The input data $x_{adv}^{k-1}$ is down-sampled by $4\times$ and is fed into the 3rd block skipping the 1st and 2nd blocks.
The perturbation $g_{ens}^{k-1}$ is first up-sampled by $4\times$ and is then added to $x_{adv}^{k-1}$ to obtain $x_{adv}^{k}$.
The \textbf{second} stage trains the top 5 blocks and the classifier. 
The input $x_{adv}^{k-1}$ is down-sampled by $2\times$ and is fed into the 2nd block skipping the 1st block.
The \textbf{third} stage trains all layers.
Note that, except for the newly added block in the second or third stage and the layers directly connected with the newly added block, all the other layers inherit the weights trained in the previous stage. 
Due to memory limitation, we set $T_t$ to a small number of 2.

The first, second, and third training stages take 100,000, 50,000, and 50,000 iterations, with the batch size of 50, 36, and 24, respectively.
Both the second and the third stages train the newly added blocks and the layers directly connected with them in the first 20,000 iterations and fine-tune all the blocks in the later 30,000 iterations.
The learning rate $\alpha$ and the number of iterations $T_t$ are set to $0.001$ and $2$, respectively.
In the first, second, and third training stages, $\epsilon_c$ is initialized to $3,000$, $1,200$, and $1,200$ respectively, and is exponentially decayed by $0.9\times$ for every $4,000$, $3,000$, and $3,000$ iterations, respectively.

We refer to the data pre-processing methods in the repository\footnote{https://github.com/tensorflow/models/tree/master/research/slim/preprocessing} on GitHub to pre-process the data used in our experiments on ImageNet.
When the resolution of the source model is 224$\times$224, we refer to `vgg\_preprocessing.py' while when the resolution is 299$\times$299, we refer to 
`inception\_preprocessing.py'.

\subsection{The network architecture}
\begin{figure*}[]
	\centering
	\includegraphics[width=0.95\textwidth]{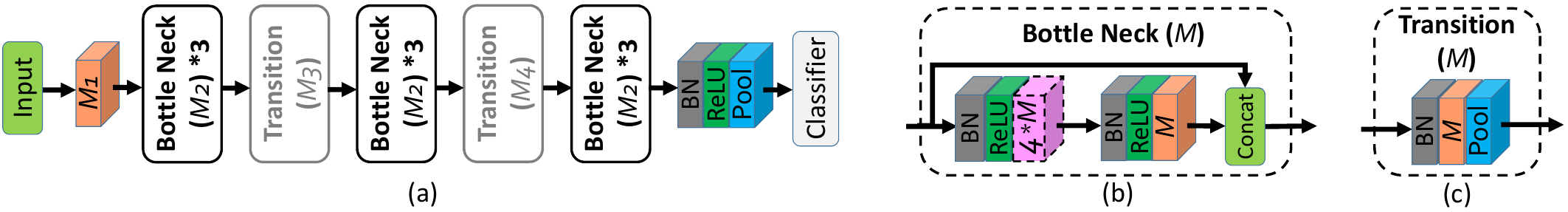}
	\caption{
		(a) DenseNet-22-BC.
		Orange cube is convolution layer with 3$\times$3 kernel size.
		Pink cube is convolution layer with 1$\times$1 kernel size.
		`Bottle Neck ($M_2$) *3' denotes three cascaded `Bottle Neck ($M_2$)'.
		The number (\emph{e.g.}, $M_1$, $4*M$, $M$) on each convolution layer denotes its number of filters.
		`Pool' in the Transition block is Max Pooling with both stride and kernel size of 2$\times$2, and the last `Pool' before the classifier is Global Average Pooling.
		(b) The detailed structure of Bottle Neck.
		(c) The detailed structure of Transition.
	}
	\label{fig:network_dense}
\end{figure*}

\textbf{DenseNet-22BC} is shown in Figure \ref{fig:network_dense}.
$M_1$, $M_2$, $M_3$, and $M_4$ are set to 80, 40, 100, and 110, respectively.
We denote MTA-PGD with DenseNet-22BC backbone as MTA-PGD$_{dense}$ and show its performances in Table \ref{tab:cifar}.

\begin{figure}[]
	\centering
	\includegraphics[width=0.49\textwidth]{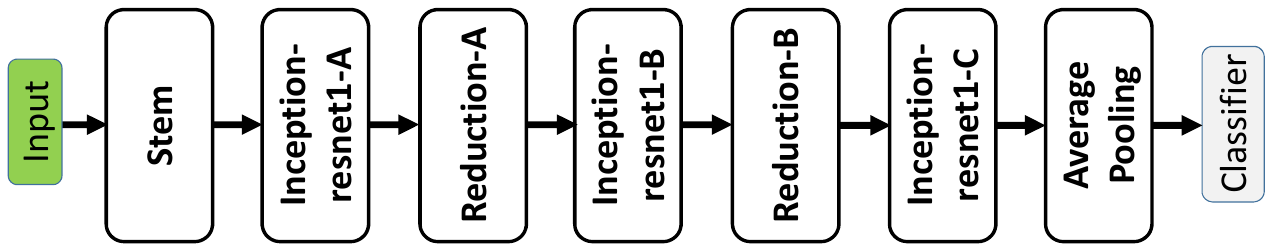}
	\caption{
		The simplified Inception network.
		All the blocks have the same inner structures with those of Inception-ResNet-V2. 
	}
	\label{fig:network_Inc}
\end{figure}

\textbf{The simplified Inception network} is a much shallower and thinner version of the official Inception-ResNet-V2.
Figure \ref{fig:network_Inc} shows the structure of the simplified Inception.
The official Inception-ResNet-V2 repeats each Indeption-resnet1-A, -B, or -C block for several times while the simplified Inception does not repeat them.
We denote MTA-PGD with this backbone as MTA-PGD$_{Inc}$ and show its performances in Table \ref{tab:imagenet_one}.

\subsection{Implementations of the compared methods.}
For fair comparisons between MTA-PGD and the compared methods, we tune the compared methods for their best possible performances in our re-implementation.
$\epsilon$ is set to 15 by default for all methods and $T_v$ is set to 10 for all PGD-based methods.

\textbf{M-PGD} utilizes gradient momentum to make the generated adversarial examples more transferable.
The most important hyper-parameter of M-PGD is $\mu$.
In our implementation, we found that setting $\mu$ to 1 can achieve the best transfer attack performance.

\begin{figure*}[t]
	\centering
	\includegraphics[width=0.8\textwidth]{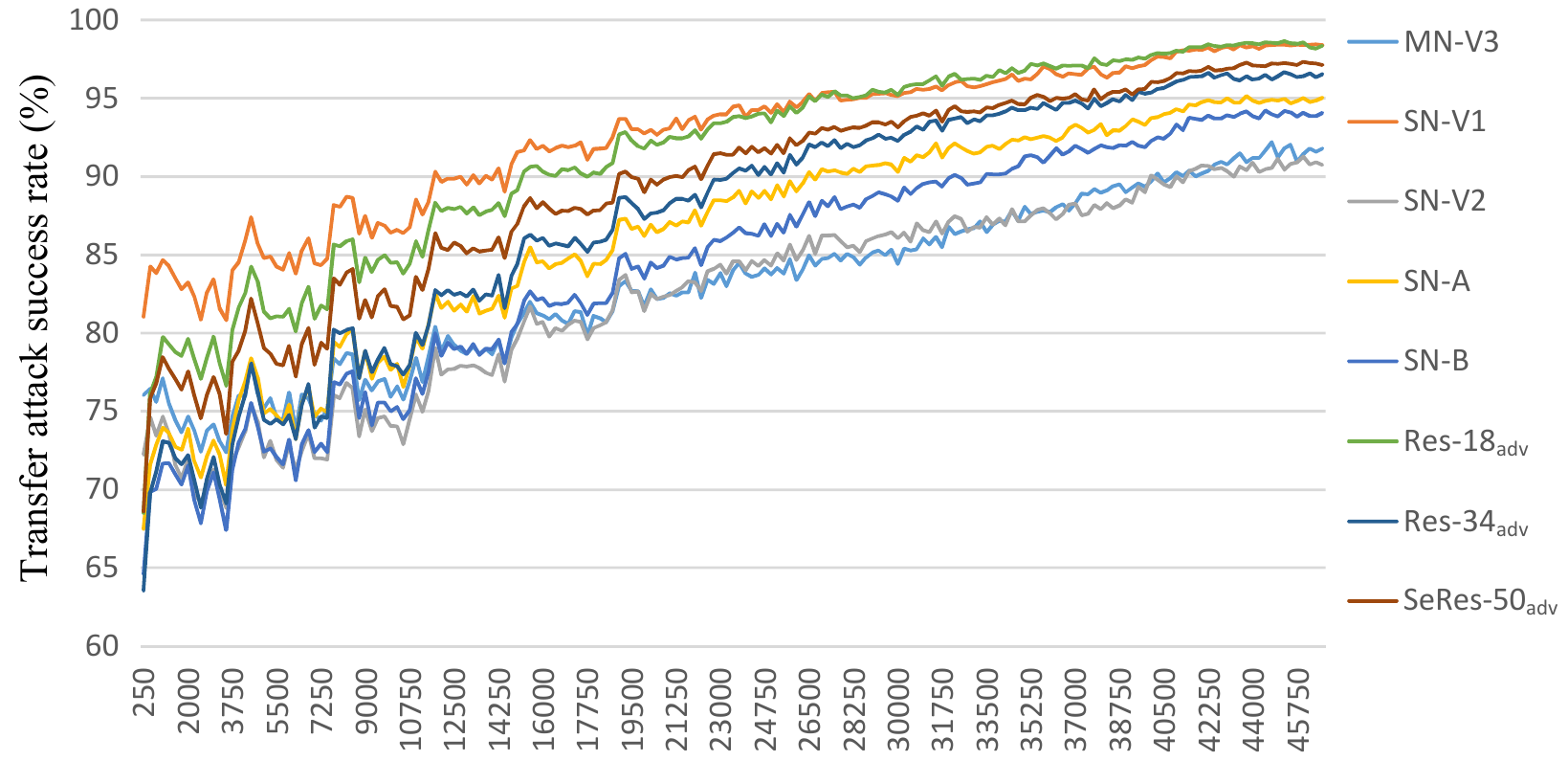}
	\caption{
		Transfer attack success rates of MTA-PGD on the eight black-box Cifar-10 models, across the training process.
	}
	\label{fig:curves_cifar}
\end{figure*}

\textbf{{DI-PGD}}.
We follow the available public code\footnote{https://github.com/cihangxie/DI-2-FGSM\label{code}} of DI-PGD to implement it in our experiment.
As to the experiments on ImageNet, we set `FLAGS.image\_width' and `FLAGS.image\_resize' (two parameters of the input\_diversity function in the official code\textsuperscript{\ref{code}}) to 224 and 256 respectively.
On Cifar-10, we set `FLAGS.image\_width' and `FLAGS.image\_resize' to 32 and 36, respectively.
For all experiments, we set $p$ to 0.8.

\textbf{TI-PGD}.
We utilize the public code\footnote{https://github.com/dongyp13/Translation-Invariant-Attacks} of TI-PGD to implement it in our experiment.

\textbf{SGM-PGD} uses a parameter $\gamma$ to reduce the gradient from all residual modules of ResNet or DenseNet.
We utilize grid search to tune $\gamma$ for each ResNet and DenseNet source model shown in Table \ref{tab:imagenet_multi}.
We denote $\gamma$ for the source model of Res-50, Res-152, DN-161, and DN-121 as $\gamma_{res50}$, $\gamma_{res152}$, $\gamma_{dn161}$, and $\gamma_{dn121}$, respectively.
The tuned best $\gamma_{res50}$, $\gamma_{res152}$, and $\gamma_{dense}$ for the source model group Res-50+Res-152+DN-161 are 0.20, 0.45, and 0.70, respectively.
The tuned best $\gamma_{res50}$ and $\gamma_{dn121}$ for the source model group Res-50+Inc-V1+DN-121 are 0.60 and 0.85, respectively.
The tuned best $\gamma_{res50}$ for the source model group Res-50+Inc-V1 is 0.65.

\textbf{A-PGD}. We utilize the public public code\footnote{https://github.com/fra31/auto-attack} of A-PGD to implement it in Tables \ref{tab:cifar}.

\textbf{SI-N-PGD}.
We refer the official code \footnote{https://github.com/JHL-HUST/SI-NI-FGSM} to implement SI-N-PGD in Table \ref{tab:imagenet_one}.
The number of scale copies is set to $m$ = 5.

\textbf{AEG}. By referring to the AEG's paper and code\footnote{https://github.com/joeybose/Adversarial-Example-Games}, we re-implement AEG on Cifar-10 and train the generator and the critic for 500 epochs with the learning rate of 0.001.
The architecture of the generator is the encoder-decoder defined in Tab.7 of AEG's paper.
We do not implement AEG on ImageNet because training the generator and critic is expensive on ImageNet.

\textbf{FIA}.
We utilize the public code\footnote{https://github.com/hcguoO0/FIA} of FIA to implement it on ImageNet.
The used intermediate feature layers are the same with the description in the paper of FIA \cite{wang2021feature}.

\textbf{IR-PGD}.
We directly utilize the public code\footnote{https://github.com/xherdan76/A-Unified-Approach-to-Interpreting-and-Boosting-Adversarial-Transferability} of IR-PGD to implement it on ImageNet.
When implementing IR-PGD on Cifar-10, we set the hyper-parameter `args.grid\_scale' to 1.

\subsection{Training curves}
In the training process of the MSM, we evaluate MTA-PGD's transfer attack performances on the target models for every 250 iterations.
Figure \ref{fig:curves_cifar} visualizes the performance curves on eight Cifar-10 target models.
It is observed that with the training going on, the transfer attack success rates on the target models rise gradually.
The periodic fluctuations of the performances are caused by the periodic decay of the hyper-parameter $\epsilon_c$ described in Section \ref{sec:exp_set_cifar}.

\subsection{Training cost}
We conduct all experiments on Tesla P40 GPU.
The training cost of the MSM depends mainly on its backbone, the used source models, the dataset, the batch size, $T_t$, and \emph{etc.}.
On Cifar-10, the default backbone of the MSM is ResNet-13, the batch size is 64, $T_t=7$, and we use 8 source models to train the MSM.
The training costs one P40 GPU and approximately 2.5T FLOPs per iteration.
On ImageNet, the default backbone of the MSM is ResNet-19, $T_t=2$, the batch size is 24 in the third training stage.
When using the Inc-V3 source model to train the MSM, the third training stage costs one P40 and approximately 3.2T FLOPs per iteration.
When using the Res-152, Res-50, and DN-161 source models to train the MSM, the third training stage costs three P40 GPUs and approximately 6.5T FLOPs per iteration.

\subsection{Visualization of adversarial examples}
\begin{figure*}[]
	\centering
	\includegraphics[width=0.9\textwidth]{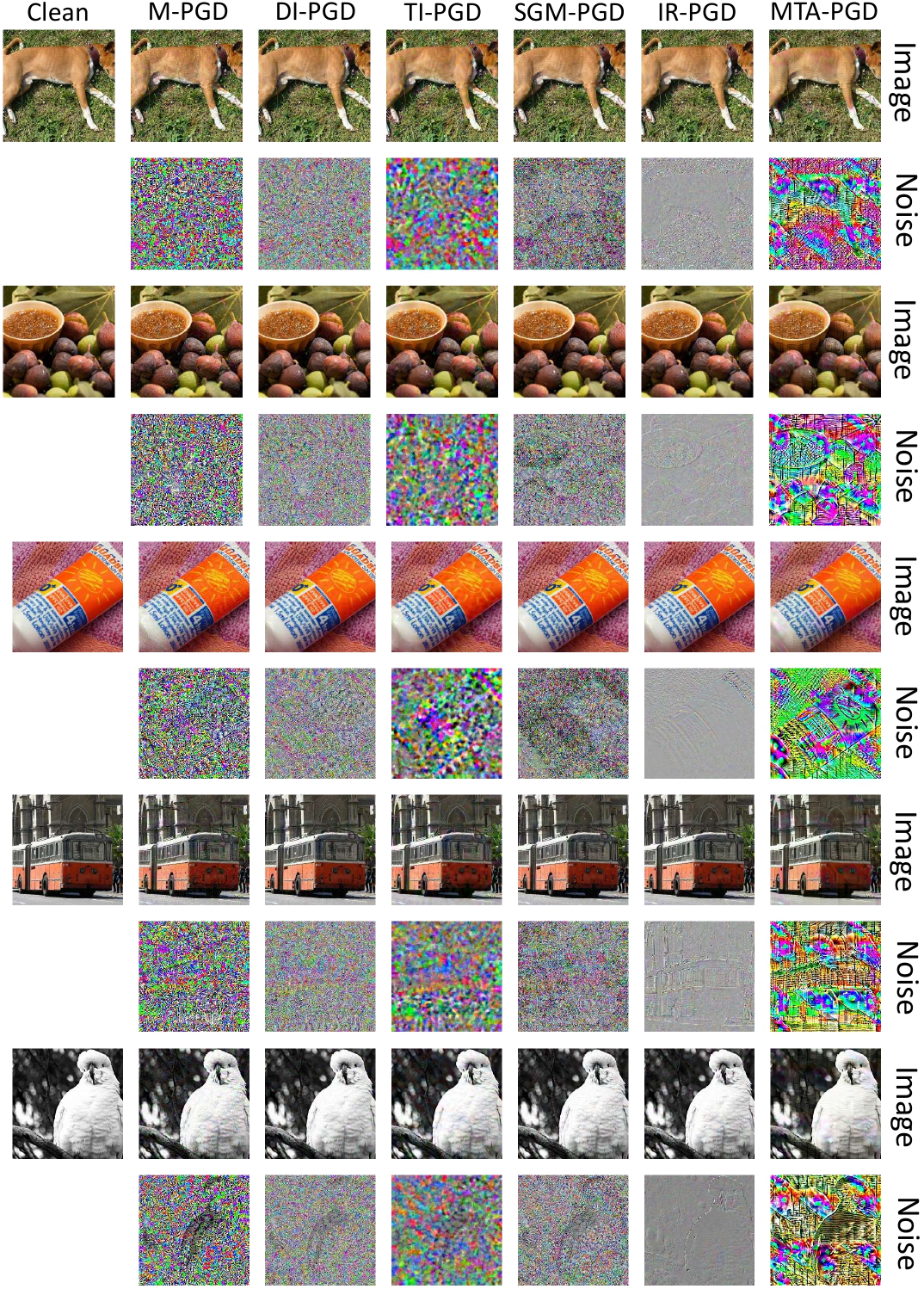}
	\caption{
		The adversarial examples and the noises generated via M-PGD, DI-PGD, TI-PGD, SGM-PGD, IR-PGD, and MTA-PGD.
		The corresponding clean images are shown in the left most column.
		The source model is Res-152.
	}
	\label{fig:adv_images}
\end{figure*}

Figure \ref{fig:adv_images} visualizes the adversarial examples and the noises generated for the corresponding clean images via M-PGD, DI-PGD, TI-PGD, SGM-PGD, IR-PGD, and MTA-PGD.
All the clean images are sampled from the validation set of ImageNet.

\end{document}